\documentclass[journal]{IEEEtran}

\usepackage[utf8]{inputenc}

\usepackage{adjustbox}
\usepackage{amsmath}
\usepackage{amssymb}
\usepackage{pifont}
\usepackage[ruled,vlined]{algorithm2e}
\usepackage{amsfonts}
\newcommand{\cmark}{\ding{51}}
\newcommand{\xmark}{\ding{55}}
\usepackage{graphicx}
\usepackage{rotating}
\usepackage{array, booktabs}
\usepackage{float,nonfloat}
\usepackage{authblk}
\usepackage{etoolbox}

\usepackage{makecell}
\usepackage[pagebackref=false,breaklinks=true,colorlinks,bookmarks=false]{hyperref}
\usepackage{multirow}
\usepackage{amsmath}
\usepackage{arydshln}
\usepackage{caption}[=v1]
\DeclareMathOperator*{\argmax}{arg\,max}
\DeclareMathOperator*{\argmin}{arg\,min}

\newcommand{\iwidth}{.1\columnwidth}
\newcommand{\twidth}{.1\columnwidth}
\newcommand{\im}[1]{ \includegraphics[width=\iwidth]{figures/dataset_examples/#1}}

\newcolumntype{C}{ >{\centering\arraybackslash} m{\iwidth}}
\newcolumntype{D}{ >{\centering\arraybackslash} m{\twidth}}

\makeatletter
\apptocmd{\@maketitle}{\centering \url{https://github.com/MartinPernus/FICE}}
\makeatother

\begin{document}

\title{FICE: Text--Conditioned Fashion Image Editing With Guided GAN Inversion}

\author[1]{Martin~Pernuš,~\IEEEmembership{Student Member,~IEEE}}
\author[2]{Clinton~Fookes,~\IEEEmembership{Senior Member,~IEEE}}
\author[1]{Vitomir~Štruc,~\IEEEmembership{Senior Member,~IEEE}}
\author[1]{Simon~Dobrišek,~\IEEEmembership{Member,~IEEE}} 
\affil[1]{University of Ljubljana}
\affil[2]{Queensland University of Technology}

\maketitle

\begin{abstract}
Fashion--image editing represents a challenging computer vision task, where the goal is to incorporate selected apparel into a given input image. Most existing techniques deal with this task by first selecting an example image of the desired apparel and then transferring the clothing onto the target person. Such techniques are commonly referred to as Virtual Try-On methods. Conversely, in this paper, we consider editing fashion images with text descriptions. Such an approach has several advantages over example--based virtual try-on techniques, e.g.: $(i)$ it does not require an image of the target fashion item, and $(ii)$ it allows the expression of a wide variety of visual concepts through the use of natural language. Existing image--editing methods that work with language--based inputs are heavily constrained by their requirement for training sets with rich attribute annotations or they are only able to handle simple text descriptions. To address these constraints, we propose a novel text--conditioned editing model, called FICE (Fashion Image CLIP Editing), capable of handling a wide variety of semantically diverse text descriptions to guide the editing procedure. 
Specifically with FICE, we augment the common GAN inversion process by including semantic, pose-related, and image-level constraints when generating the desired images. We leverage the capabilities of the pretrained CLIP model to enforce the targeted semantics, due to its impressive image--text association capabilities. We furthermore propose a latent--code regularization technique that provides the means to better control the fidelity of the synthesized images and ensures that images are generated from latent codes coming from a well-defined part of the latent space. We validate FICE through comprehensive experiments on a combination of VITON images and Fashion-Gen text descriptions and in comparison with several state-of-the-art text--conditioned image editing approaches. Experimental results demonstrate FICE generates highly realistic fashion images and leads to stronger editing performance than existing competing approaches. 

\end{abstract}

\section{Introduction}
\IEEEPARstart{F}{ashion}--image editing refers to the task of changing the appearance of a person in a given image by incorporating a desired fashion item (e.g., different apparel) in a realistic and visually convincing manner. Successful applications of such algorithms enable users to visualize and virtually try-on selected clothing from the comfort of their homes. This functionality has the potential to enable easier online apparel sales, reduce costs for retailers, and reduce the environmental footprint of the fashion industry by minimizing returns~\cite{fashion_industry}. As a result, significant research efforts have been directed towards fashion--image manipulation (or Virtual Try-On -- VTON) techniques that deliver convincing photorealistic editing results over the years~\cite{viton, cp-vton, vtnfp, yang2020towards, fele2022c, mg-vton, cyclefashion, wuton, pf-afn}.

\begin{figure}[t]
	\centering
\includegraphics[width=\columnwidth]{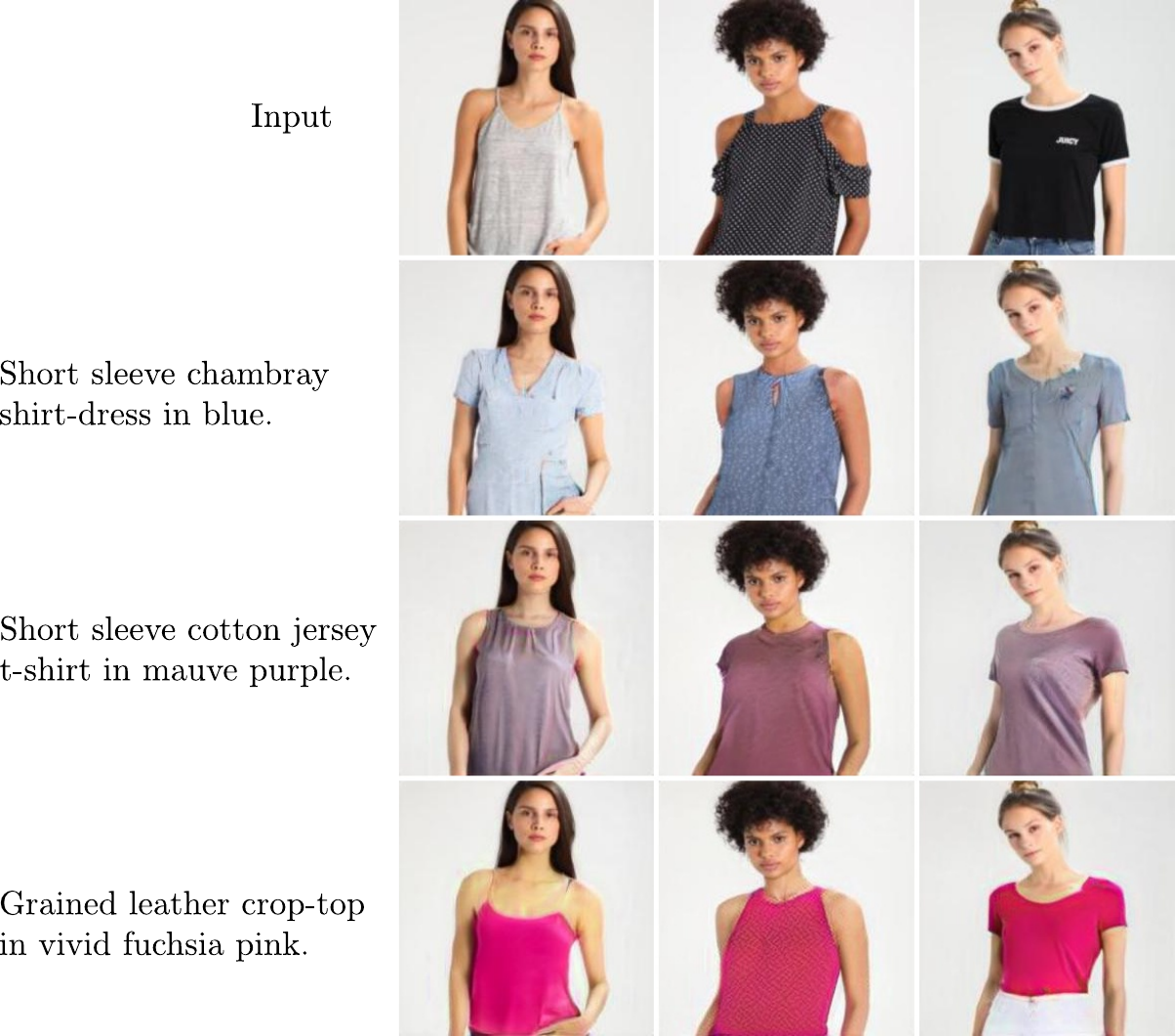}
	\caption{\textbf{Fashion--image editing with language--based inputs.} In this paper, we propose FICE (\textbf{F}ashion \textbf{I}mage \textbf{C}LIP \textbf{E}diting), a text--conditioned image editing model, capable of handling a wide variety of text inputs with the goal of manipulating fashion images toward the desired target appearance.}
	\label{fig:cherry}
\end{figure}

Recent Virtual Try-On solutions have achieved great success in synthesizing photorealistic fashion images by building on advances made in convolutional neural networks and adversarial training objectives~\cite{gan}. Most existing techniques in this area condition their editing models on example images of the target clothing, which is typically warped and stitched onto the given input image. Considerably less attention has been given to text--conditioned fashion--image editing, despite the fact that such methods represent an attractive alternative to example--based editing techniques and allow the manipulation of images through more natural high-level language--based descriptions of the desired apparel. 
While, to the best of our knowledge, only a modest amount of work has been conducted on this topic so far, existing text--conditioned methods are commonly limited to very basic descriptions, mostly due to the small size of suitable training datasets that are publicly available~\cite{prada}. To mitigate these problems, some text--conditioned fashion works proposed to parse the input text into closed sets of categories~\cite{text2human} for easier text processing, simplifying the task to a more basic, categorical problem.

Meanwhile, various image--text association models have emerged. These models are trained on hundreds of millions of image--text pairs~\cite{clip} and represent powerful tools for associating visual data and language descriptions~\cite{slip, dalle, dalle2, ju2022prompting}. 
As a result, they have been successfully deployed for text--conditioned image editing in combination with recent state-of-the-art generative adversarial networks (GANs)~\cite{styleclip}. Such solutions typically first embed the given input image into the latent space of a pretrained GAN model through a process referred to as GAN inversion~\cite{gan_inversion_survey}, and  
then perform text--conditioned manipulations in the latent space that eventually lead to semantically meaningful changes in the corresponding output images~\cite{styleclip, interfacegan, abdal2021styleflow, harkonen2020ganspace}.
While general--purpose text--conditioned GAN--based editing techniques have shown success in various settings, a straightforward application to the fashion domain is challenging and, more importantly, does not guarantee optimal editing results. This is due to the inherent reconstruction--editability trade--off~\cite{e4e} of such techniques, which typically result in a significant loss of identity information as well as pose changes when inverting an image into a GAN latent code. Furthermore, despite the recent advances in disentangled editing in the GAN latent space~\cite{styleclip, s-space}, such methods are still problematic to use in the context of text--conditioned editing due to the high sensitivity to hyperparameter choices~\cite{maskfacegan}.   

In this paper, we address these open challenges through the introduction of FICE (Fashion Image CLIP Editing) -- a novel text--conditioned image--editing approach tailored towards fashion images. FICE builds on the standard GAN inversion framework, but extends the GAN inversion process to allow for the integration of novel capabilities that enable text--conditioned fashion image editing. In contrast to works that edit fashion images with the help of categorical attributes~\cite{prada}, FICE enables fashion image editing through text as the only (semantics--related) conditioning signal. To facilitate editing of fashion images with FICE, we propose an iterative GAN inversion procedure that utilizes several constraints, when optimizing for the latent code with the desired target semantics, i.e.: $(i)$ a \textit{pose--preservation constraint} that ensures that the pose of the subject in the image is not altered during the editing process, $(ii)$ a \textit{composition constraint} that uses a segmentation model (i.e., a body parser) to identify regions (head and garment areas) in the input image to preserve and/or alter, and  $(iii)$ a \textit{semantic--content constraint} that enforces the semantics expressed in the provided text descriptions. We use various differentiable deep learning models to implement the constraints and leverage the CLIP model, a recent state-of-the-art image--text association approach, to enforce the desired semantics. Furthermore, we propose a latent--code regularization objective to ensure more realistic editing results. Finally, we also utilize an image--stitching step to combine relevant image regions from the original and edited images in the final overall result.

To demonstrate the capabilities of FICE, we perform rigorous experiments on images from the VITON image dataset~\cite{viton}, combined with text descriptions from the Fashion-Gen dataset~\cite{fashiongen}. We compare FICE to several general text--conditioned GAN--based editing methods and show that the proposed approach leads to superior editing results for fashion images. A few of these results can be seen in Fig.~\ref{fig:cherry} for three different text descriptions. Our research leads to the following main contributions that are presented in this paper:

\begin{itemize} 
	\item We propose FICE, a GAN--inversion based approach for text--conditioned fashion image editing, which can be used with a wide variety of textual inputs and leads to realistic and visually convincing editing results.
	\item We introduce a regularization technique for the GAN inversion procedure to minimize the generation of images outside the GAN learned distribution.
	\item Through quantitative and qualitative evaluations, we show the benefits of text--based  editing of fashion images and demonstrate that FICE convincingly outperforms competing (state-of-the-art) text--based editing techniques.
\end{itemize}

\section{Related Work}

In this section, we review relevant prior work and discuss existing research on $(i)$ generative adversarial networks, $(ii)$ text--conditioned image generation and editing, $(iii)$ GAN inversion techniques, and $(iv)$ the use of computer vision in fashion. The goal of the section is to provide the necessary background for our work. A more comprehensive coverage of these topics can be found in some of the recent surveys, e.g.,~\cite{gan_inversion_survey, wu2017survey, cheng2021fashion}.  

\subsection{Generative Adversarial Networks}
	Generative Adversarial Networks (GANs)~\cite{gan} have in recent years become the \textit{de-facto} method for  unconditional image synthesis, allowing convincing high resolution image synthesis and reasonable training times with consumer-grade hardware. DCGAN~\cite{dcgan} introduced convolutional GANs and provided architectural pointers to achieve successful GAN convergence. ProGAN~\cite{progan} was the first GAN model that achieved megapixel-sized images thanks to a progressive lear\-ning scheme.
    StyleGAN~\cite{stylegan1} introduced a non-linear mapping of the latent space and an alternative generator design, inspired by the style transfer literature. 
 StyleGAN2~\cite{stylegan2} further adjusted the generator architecture to remove the frequent droplet artefacts and regularized the training with path-length regularization. StyleGAN2-ADA~\cite{stylegan2-ada} proposed several augmentation techniques to enable learning a high-quality GAN with limited training data. StyleGAN3~\cite{stylegan3} presented a continuous interpretation of the generator signals to prevent the dependence of the generated image on the absolute pixel coordinates, in turn, enabling more natural latent code interpolations.

Next to the architectural advances, considerable progress has also been made in the field of GAN regularization and training. Various techniques of stabilizing the GAN convergence by modifying the training objective were proposed in~\cite{wgan, wgan-gp, lsgan}. Other successful methods include weight normalization~\cite{miyato2018spectral} and various regularization approaches~\cite{mescheder2018training}. 

\subsection{Text--Conditioned Image Generation and Editing}

Text--conditioned image generation models are focused on generating realistic images that match the semantics of the provided text descriptions. Conversely, corresponding \textit{editing techniques} try to realistically manipulate images in a way that preserves the image characteristics irrelevant to the text description. Thus, text--conditioned image editing aims to alter only the semantic content that is expressed in the text description, while preserving all other parts of the data. 

\textbf{Image Generation.} The seminal work of Reed \textit{et al.}~\cite{reed2016generative} proposed a text--conditioned GAN model by feeding the text information to both, the generator and discriminator of the GAN design. StackGAN~\cite{stackgan} and StackGAN++~\cite{stackgan++} proposed stacked generators, where the resolution of the generated images increased progressively with each generator in the stack. AttnGAN~\cite{attngan} proposed an attention mechanism to attend to relevant words, on which to  condition the image--generation process. MirrorGAN~\cite{mirrorgan} proposed a cyclic GAN architecture that regularized the generated image by enforcing correct (re)descriptions. 

While these early models discussed above already provided for competitive performance, more recent text--conditioned generative models are several orders of magnitude larger in size and are trained on several orders of magnitude larger datasets. DALL-E~\cite{dalle}, for example, trained a $12$-billion parameter autoregressive transformer on $250$ millions of image--text pairs, and considerably outperformed previous models in the considered (zero--shot) evaluation experiments. The model was further improved in DALL-E 2~\cite{dalle2} through the use of diffusion techniques. Imagen~\cite{imagen} proposed to increase the size of the text encoder to achieve better results in terms of image fidelity and image--text alignment.

\textbf{Image Editing.} An early approach to text--conditioned image editing methods was described by Dong \textit{et al.} in~\cite{dong2017semantic}. Here, the authors proposed a conditional encoder-decoder GAN model and reported impressive results on two diverse datasets. Nam \textit{et al.}~\cite{nam2018text} proposed so-called word-subset local discriminators that enabled fine-grained image editing. ManiGAN~\cite{li2020manigan} proposed a different strategy for merging image and text representations while adding a detail correction module for enhanced image quality. 

Another notable group of methods performs image editing by first converting a given image into the latent
code of some pre-trained GAN, in a process known as GAN inversion, then performing various latent code manipulations to achieve the desired edits. InterFaceGAN~\cite{interfacegan}, for example, identified directions in the latent space of StyleGAN2 that corresponded to specific semantic changes (given by binary attribute labels) in the corresponding output image. Image2StyleGAN~\cite{image2stylegan, image2stylegan++} performed several face image edits using GAN inversion and StyleCLIP~\cite{styleclip} proposed different methods for text-guided image editing, where the general idea is based on combining
the generative capabilities of StyleGAN with the image--text matching capabilities of the CLIP model~\cite{clip}. TediGAN~\cite{tedigan} introduced a control mechanism based on style mixing in StyleGAN to achieve the desired semantics in face images driven by text descriptions. 

\subsection{GAN Inversion} 

As can be seen from the literature review presented above,  a considerable amount of existing editing techniques deals with the process of GAN inversion to retrieve the image's latent code for editing. How to conduct the GAN inversion is a key consideration with these techniques that has a significant impact on the final editing capabilities. 
Richardson \textit{et al.}~\cite{psp}, for example, proposed an encoder, called pSp, for projecting images into the StyleGAN latent space, and demonstrated its feasibility through several image-to-image translation tasks. E4e~\cite{e4e} adjusted the pSp model so that the latent codes follow a similar distribution as the original StyleGAN latent codes and performed image edits with several latent code manipulation techniques.
ReStyle~\cite{restyle} presented an iterative procedure to obtain the latent code, while HyperStyle~\cite{hyperstyle} adjusted the StyleGAN generator weights on a per-sample basis to achieve better image reconstructions.
Additional GAN inversion methods can be found in the recent survey~\cite{gan_inversion_survey}.

Similarly to the techniques discussed above, the  proposed FICE model also relies on GAN inversion to perform fashion image editing. However, different from alternative techniques, our model is geared towards the characteristics of fashion images and exploits an iterative inversion process that explicitly considers pose preservation and image stitching constraints in addition to the targeted semantics to ensure both the desired garment appearance as well as identity preservation.

\subsection{Computer Vision in Fashion} 
A considerable cross-section of fashion-related computer-vision research focuses on Virtual Try-On technology, where, given an image of a person and some target garment, the goal is to realistically fit the garment, while preserving the original person pose and appearance.
VITON~\cite{viton}, CP-VTON~\cite{cp-vton} proposed to warp the target clothing by conditioning the warping procedure on a coarse human shape and pose map before blending it with the input image. VTNFP~\cite{vtnfp} and C-VTON~\cite{fele2022c} adopted semantic segmentation models to guide the synthesis. Yang \textit{et al.}~\cite{yang2020towards} constrained the fashion item warping and introduced adaptive content generation and preservation constraints. MG-VTON~\cite{mg-vton} proposed a network that enabled multi--pose try-on. VITON-HD described a
Virtual Try-On method for higher resolution image generation. DCTON~\cite{cyclefashion} utilized cycle--consistency for the editing procedure~\cite{cycle}, and WUTON~\cite{wuton} and PF-AFN~\cite{pf-afn} proposed a teacher--student setup that removed the need for intermediate auxiliary  models (e.g., for parsing, pose-estimation) during the editing step.

\begin{figure*}[ht]
	\centering
	\includegraphics[width=\textwidth]{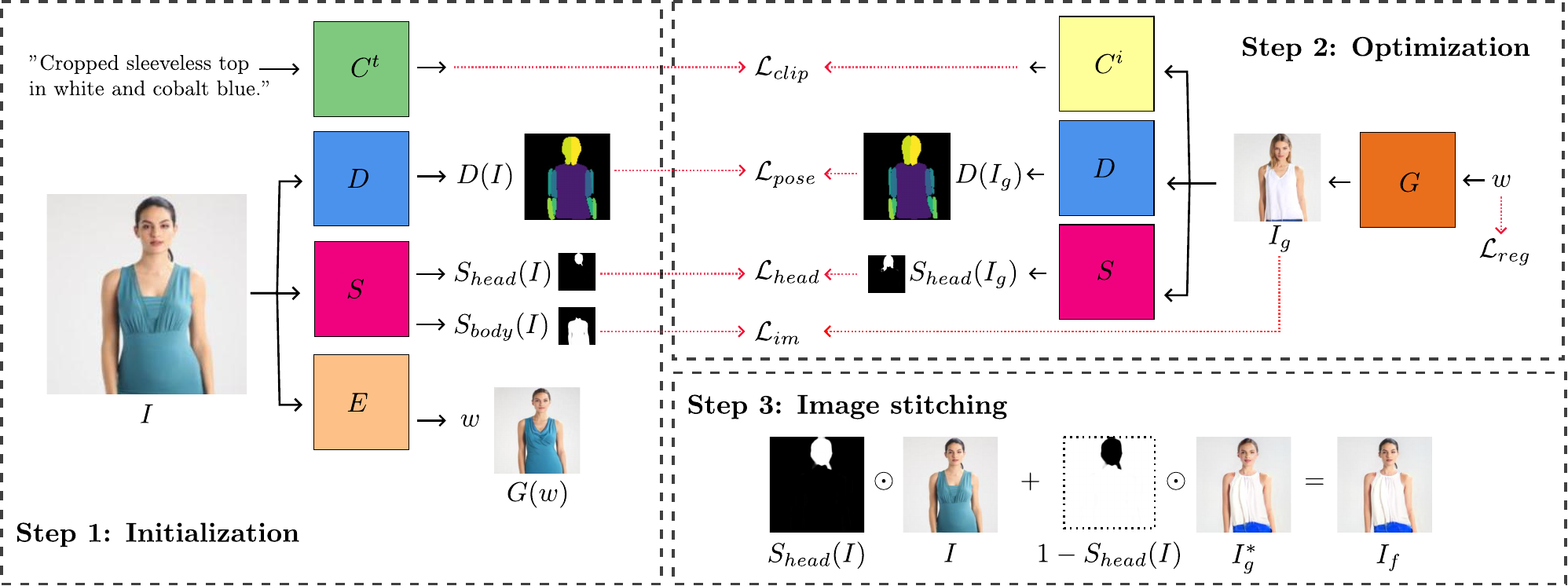}
    \caption{\textbf{High-level overview of the proposed FICE editing model}. The model uses a constrained GAN inversion procedure over the latent space of the pretrained StyleGANv2 model $G$ to incorporate the semantics expressed in the provided text description into the output image $I_g$. Here, the optimization--based GAN inversion is performed by enforcing: $(i)$ the presence of semantic content (determined by the image--text association model $C$), $(ii)$ pose preservation (ensured by the pose parser $D$), $(iii)$ image consistency (through the segmentation model $S$), and $(iv)$ latent code regularization. An image stitching step is also incorporated into FICE to improve the image fidelity and further enhance the model's identity preservation capabilities.\label{fig:scheme} 
    }
\end{figure*}

Work has also been performed to enable text conditioning for editing or generating target images. FashionGAN~\cite{prada}, for example, edited images conditioned on text inputs, segmentation masks and several image--specific categorical attributes using encoder--decoder GANs. Fashion-Gen~\cite{fashiongen} introduced a dataset of image--text pairs and experimented with unconditional and text--conditioned image generation. Recently, Text2Human~\cite{text2human} proposed generating human images based on a description of clothes shape and texture. With this approach, the text--encoding method mapped the input text into a number of closed sets of categories, which limits the language expressiveness of the input text.    

Unlike the methods presented above, FICE uses text as the only condition/input for processing fashion images. Furthermore, the model is not focused on image generation but rather image editing, i.e. on redressing persons, similarly as Virtual Try-On methods, but using only text descriptions as input instead of image examples. Finally, unlike some methods, that parse the text into a closed set of categorical attributes, FICE relies on CLIP~\cite{clip}, an image--text association model, trained on $400$ million image--text pairs, that allows the proposed approach to process a wider set of linguistic concepts than alternative solutions from the literature. 

\section{Methodology}

The main contribution of this paper is a novel text--conditioned model for (fashion) image editing, named FICE (\textbf{F}ashion \textbf{I}mage \textbf{C}LIP \textbf{E}diting). In this section, we provide an in-depth description of the proposed model and elaborate on its characteristics.

\subsection{Problem Formulation and FICE Overview}

The aim of FICE is to edit the given (fashion) image $I \in \mathcal{R}^{3 \times n \times n}$ in accordance with some (appearance--related) text description $t$ and to synthesize a corresponding output image  $I_f \in \mathcal{R}^{3 \times n \times n}$ that adheres as closely as possible to the semantics expressed in $t$. Here, the synthesis process needs to meet the following criteria: (1) the synthesized output image $I_f$ should preserve the pose, identity and other appearance characteristics of the subject in $I$, (2) the editing process should be local and only affect the desired fashion items (e.g., apparel), while leaving other parts of $I$ unchanged,  (3) clothing appearances, encoded in $t$, need to be realistically and seamlessly integrated into $I_f$ taking the initial pose and body shape into account, and (4) a wide variety of textures and clothing designs need to be supported. Thus, the goal of FICE is to implement an image-to-image mapping $\psi_t$ conditioned on $t$, i.e. 
\begin{equation}
	\psi_t: I \rightarrow I_f \in \mathcal{R}^{3 \times n \times n},
\end{equation}
under the constraints discussed above. As illustrated in Fig.~\ref{fig:scheme}, FICE defines the mapping through a three--stage procedure that consists of: $(i)$ an \textit{Initialization} stage, $(ii)$ a \textit{Constrained GAN--inversion} stage, and  $(iii)$ an \textit{Image Stitching} stage. A high-level summary of the three (key) steps is given below:  
\begin{itemize}
	\item \textbf{Initialization.} In the first step, the model  initializes a latent code $w$ based on the image input $I$ using a GAN inversion encoder $E$, which serves as a first approximation of the targeted latent code. This initial code approximates the original appearance of the input image, when interpreted through the pretrained GAN generator $G$, i.e., $I \approx G(w)$, as shown on the left of  Fig.~\ref{fig:scheme}. Additionally, a dense-pose representation and segmentation masks corresponding to different body parts are also computed from $I$ during this step. 
	\item \textbf{Constrained GAN Inversion.} Next, the initial latent code $w$ is further optimized using a (constrained) optimization--based GAN inversion technique. At each step of the optimization procedure, the latent code $w$ is fed to the generator $G$ to synthesize an intermediate image $I_g = G(w)$. An optimization objective (loss) $\mathcal{L}$ is then defined over $I_g$ to drive the GAN inversion that ensures that: $(i)$ the semantics defined in $t$ are present in $I_g$, $(ii)$ the pose in $I$ and $I_g$ are the same, $(iii)$ the editing procedure appears natural, and that $(iv)$ the optimized latent code lies in a well-defined part of the GAN latent space. The result of the (fixed-step) optimization procedure is a latent code $w^*$,
	\begin{equation}
	    w^* =\argmin_w \{\mathcal{L}(I,G(w),t)\},
	    \label{eq: contrainedGANinv}
	\end{equation}
	that corresponds to the final optimized output image $I_g^* =G(w^*)$. A number of (auxiliary) differentiable models are utilized to facilitate the optimization procedure -- see Step 2 in Fig.~\ref{fig:scheme}. Details on the models are provided in the following sections.   
	\item \textbf{Image Stitching.} In the final stage, a simple image--composition process is utilized to combine the (optimized) GAN synthesized/optimized image $I_g^*$ with the original image $I$ to ensure identity preservation and to produce the final output image $I_f$.
\end{itemize}

\subsection{FICE Structure}

FICE relies on several (differentiable) models to infer information on the high-level characteristics of the given input image for the constrained GAN inversion procedure:
\begin{itemize}
	\item \textbf{Generator} ($G$). The key component of FICE is the (pretrained) GAN generator $G$, which is responsible for synthesizing the fashion images based on the provided latent code $w$ and defines the characteristics of the generated data. To this end, we selected the StyleGANv2 model from~\cite{stylegan2}, which is capable of generating high-quality images of a resolution up to $1024\times1024$ px. 
	\item \textbf{CLIP} ($C$). The Contrastive Language--Image Pre-training model (or CLIP for short)~\cite{clip} is a neural network trained for pairing images and text. It consists of separate image and text encoders that produce embeddings for the pairing. The model has been trained on $400$ million (image, text) pairs and was demonstrated to be suitable for zero--shot transfer to various downstream tasks. Inspired by these zero--shot capabilities,  we use CLIP to provide semantic knowledge to our model and edit the image according to the provided text description $t$.
	\item \textbf{DensePose} ($D$). A critical aspect for realistic results, when editing fashion images, is pose preservation. 
	To preserve the complete body structure and pose information, we employ a sophisticated parsing model, i.e., DensePose~\cite{densepose}, capable of parsing individual human body parts from the given input image. The model provides for a comprehensive and dense pose description that is utilized in FICE to ensure that the subject's pose in the original $I$ and  optimized image $I_g^*$ match.   
	\item \textbf{Segmentation Model} ($S$). A simple segmentation model based on the DeepLabv3~\cite{deeplabv3} architecture is used to identify image regions to either alter or preserve. Additionally, the goal of the model is also to ensure consistent image characteristics, so the final stitched image is artifact-free, photorealistic and visually convincing. 
	\item \textbf{Encoder} ($E$). The last model needed for image editing with FICE is a GAN inversion encoder that computes the initial latent space embedding $w$ from the input image $I$. We select the E4e (Encoder for editing~\cite{e4e}) encoder for this task for its competitive performance. 
\end{itemize}

\subsection{Constrained GAN Inversion}
The main component of FICE is a (novel) \textit{ constrained GAN inversion technique} that, given the input image $I$ and text description $t$, optimizes for the latent code $w^*$ that fits a number of predefined constraints -- see Eq.~\eqref{eq: contrainedGANinv}. After initializing the latent code $w$ and computing the corresponding output image ${I_g = G(w)}$, the goal of the optimization--based inversion process is to adjust the latent code to best match the text description while preserving various appearance characteristics of the input image. 
Thus, given a latent code $w$ and the associated image ${I_g = G(w)}$, we define the several optimization objectives for FICE, as detailed below.

\textbf{Semantic Content.}
To ensure that the (appearance-related) semantic content, expressed in the text description $t$, is present in the generated image $I_g$, we implement a CLIP--based optimization objective/loss,
\begin{equation}
	\mathcal{L}_{clip} = 1 - \cos(C^i(I_g), C^t(t)),
    \label{eq:clip}
\end{equation}
where $C^i$ and $C^t$ represent the image and text encoder of the CLIP model, respectively. The objective penalizes angular differences between the image and text embeddings, and, thus, promotes the presence of the semantics from $t$ in the generated image $I_g$, where ${I_g = G(w)}$.   

\renewcommand{\iwidth}{.32\columnwidth}
\renewcommand{\im}[1]{\includegraphics[width=\iwidth]{figures/body-parsing/#1.jpg}}
\newcommand{\imrow}[1]{\im{#1} & \im{densepose-#1} & \im{segm-#1}}
\setlength{\tabcolsep}{2pt}
\renewcommand{\arraystretch}{1.2}

\begin{figure}[t]
    \centering
    \begin{tabular}{CCC}
        \imrow{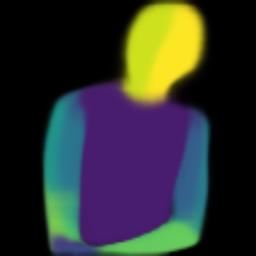} \\
        \imrow{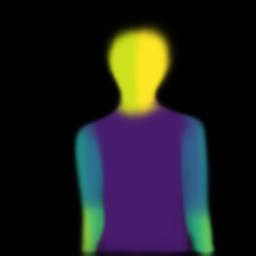} \\
        Input &  Parsed by $D$ & Parsed by $S$ \\
    \end{tabular}
    \caption{\textbf{Sample images parsed with $D$ and $S$.} The DensePose model $D$ generates a parsed body representation consisting of 24 body parts in a clothing agnostic manner. The segmentation model $S$, on the other hand, parses the input image into three classes: head (with hair), body (with clothing) and background.} 
    \label{fig:parsing}
\end{figure}

\begin{table*}[t] 
	\caption{\textbf{Overview of the fashion datasets used in the paper.} Three different datasets are selected for the experiments and used to train and test various components of FICE. \vspace{2mm}}
	\centering
    \begin{tabular}{l|rrrrrrrrr}
        \hline
        \hline
        \textbf{Dataset} & \textbf{\# Images} && \textbf{Segmentations} && \textbf{Text} && \textbf{\# Text Descriptions} && \textbf{Aim} \\
        \hline
        VITON~\cite{viton} & $16,253$ && Coarse && \xmark && n/a &&$G,E$ training/testing$^{\dagger}$ \\
        DeepFashion Retrieval~\cite{deepfashion} & $52,713$ && Fine && \xmark && n/a &&$S$ training \\
        Fashion-Gen~\cite{fashiongen} & $293,018$ && \xmark & &\cmark &&$293,018$ &&Source of text descriptions \\
        \hline
        \hline
        \multicolumn{7}{l}{\footnotesize $^{\dagger}$ Training and testing data are disjoint.}
    \end{tabular}
	\label{tab:datasets}
\end{table*}

\textbf{Pose Preservation.}
To achieve perceptually convincing editing results, it is critical to preserve the position of all body parts from the input image in the edited output. Because our goal is a detailed and accurate pose preservation, where individual body parts retain their size and shape regardless of the overlaid clothing, we utilize the powerful DensePose model $D$, which predicts the position and shape of individual body parts in a \textit{clothing-agnostic} manner. The pose--preservation constraint is, therefore, defined through the following loss, 
\begin{equation}
	\mathcal{L}_{pose} = \frac{1}{N_D} \sum_{j=1}^{N_D} ||D_j(I) - D_j(I_g)||^2_2,
    \label{eq:pose_preservation}
\end{equation}
where $N_D$ is the number of parsed body parts, $D$ is the pose parsing model and $D_j(\cdot)$ denotes the parsed mask of the $j$-th body part. A couple of example outputs produced by the pose parser $D$ are shown in Fig.~\ref{fig:parsing}.

\textbf{Latent Code Regularization.} 
FICE operates in the extended latent vector space $\mathcal{W}^+$ of the pretrained StyleGAN generator, which is commonly used with (GAN--inversion) editing techniques from the literature~\cite{image2stylegan, image2stylegan++}. The latent code $w \in \mathcal{W}^+$ consists of several latent codes, each impacting an individual convolutional layer in the StyleGAN generator network $G$. The number of individual latent codes depends on the resolution of the generator network. 
However, it needs to be noted that StyleGAN's extended latent space is an artificial extension of the original latent space that, while being capable of embedding a wider variety of images, does not guarantee the synthesis of as photorealistic images as the original latent space~\cite{e4e}. 
We, therefore, propose a simple regularization mechanism for the latent codes by minimizing the distance of the coarsest latent code to the latent codes from all other layers. Concretely, given an extended latent code $w = \{w^1, w^2, ..., w^{N_w}\}$, where $N_w$ denotes the number of individual codes, we define the following loss function,
\begin{equation}
	\mathcal{L}_{reg} = \frac{1}{N_w-1} \sum_{j=2}^{N_w} ||w^1 - w^j||^2_2.
    \label{eq:regularization}
\end{equation}
The presented loss term aims at minimizing the differences between the latent codes corresponding to different convolutional layers in the StyleGAN model, which directly correlates with the minimization of the image distribution discrepancy between the vanilla and the extended latent spaces. 

\textbf{Image Composition.}
Finally, to ensure seamless image stitching, 
FICE uses an additional segmentation model $S$ that parses the following three categories from the given input image: `background' ($S_{bg}$), `body' ($S_{body}$) and `head' ($S_{head}$). Different from the pose-parsing model $D$, the components inferred by $S$ contain additional features, i.e., $S_{body}$ also captures the clothing of the subject and the $S_{head}$ also captures the hair shape, as shown in Fig.~\ref{fig:parsing}. 

We define two optimization objectives based on the segmentation model $S$. 
The first (the image loss $\mathcal{L}_{im}$) aims to preserve the background and face regions of the image. This term is necessary because although the face region is stitched with the synthesized image in the final step, this term helps to preserve the skin tone of the subject and the color 
characteristics of the input image. The loss term is defined as,
\begin{align}
	M &= 1 - S_{body}(I) \\
	\mathcal{L}_{im} &= ||M \odot (I_g-I)||^2_2,
    \label{eq:composition-image}
\end{align}
where $1\in\mathbb{R}^{n\times n}$ is a matrix of all ones and $\odot$ denotes the Hadamard product. Furthermore, we use a second loss term that preserves the `head' region,
\begin{equation}
	\mathcal{L}_{head} = ||S_{head}(I) - S_{head}(I_g)||^2_2.
    \label{eq:composition-head}
\end{equation}
This loss is used primarily to preserve the hair of the input image and to account for interactions of the hair/head and body regions that cannot be managed solely by the image loss term $\mathcal{L}_{im}$ defined above.

\textbf{Final Optimization Objective.}
The final optimization objective for the constrained GAN inversion techniques used by FICE is defined as a weighted superposition of the individual losses, as follows, 
\begin{equation}
	\begin{split}
		\mathcal{L} = &\lambda_{clip} \mathcal{L}_{clip} + \lambda_{pose} \mathcal{L}_{pose} + \lambda_{reg} \mathcal{L}_{reg} \\
		& + \lambda_{im} \mathcal{L}_{im} + \lambda_{head} \mathcal{L}_{head}, 
	\end{split}
	\label{eq:final}
\end{equation}
where $\lambda_{clip}$,$\lambda_{pose}$,$\lambda_{reg}$,$\lambda_{im}$, and $\lambda_{head}$ are balancing weights. 

\subsection{Image Stitching}
In order to preserve the identity of the input person, we perform image stitching as the final step of FICE. The final image $I_f$ is obtained as,
\begin{equation}
	I_f = S_{head}(I) \odot I + (1 - S_{head}(I)) \odot I^*_g,
\end{equation}
where $I^*_g=G(w^*)$ is the image that corresponds to the optimized latent code $w^*$ based on Eq.~\eqref{eq: contrainedGANinv}.

\section{Experimental Setup}

In this section, we describe the experimental setup used to demonstrate the capabilities of FICE. Specifically, we discuss datasets and evaluation protocols, the baseline techniques considered as well as relevant implementation details.

\subsection{Datasets}

Three types of datasets are selected for the experiments. The selected datasets, summarized in Table~\ref{tab:datasets}, provide image and text data for training and testing of the main FICE components. Example images from the three datasets are presented in Fig.~\ref{fig:dataset_examples}.

\textbf{Image Dataset.}
We use the VITON dataset as our main image database. VITON is a fashion dataset with $16,253$ frontal-view images of female models with different tops. The images are split into a training and a test set with $14,221$ and $2,032$ images, respectively. We use the training set to train the GAN model (including $G$) and E4e encoder ($E$) and the disjoint test set for performance evaluations. Similarly to related studies from the literature~\cite{image2stylegan, image2stylegan++, pulse}, we select a test set of manageable size and use the first $120$ images of the official VITON test set for the experiments.

\textbf{Segmentation Dataset.}
To train the segmentation model $S$ needed for FICE, we utilize the DeepFashion dataset~\cite{deepfashion}, specifically, the In-shop Clothes Retrieval part of the dataset. We minimize the discrepancy with the distribution of the VITON data by filtering the images to women subjects with frontal pose and restricting the fashion categories to `Blouses \& Shirts' and `Tees \& Tanks'. Furthermore, the reference segmentation masks of different classes are merged to fit the requirements of FICE, resulting in three final segmentation targets: Body, Face \& Hair and Background. The images and segmentation masks are padded to a square shape and then downscaled to a resolution of $ 256 \times 256$ px, as shown 
in Fig.~\ref{fig:parsing}.

\textbf{Text Dataset.}
The last dataset used for the experiments is Fashion-Gen~\cite{fashiongen}. We use this dataset to obtain the clothing descriptions needed for testing. Fashion-Gen consists of $293,008$ images, each paired with a corresponding text description. We perform our experiments with image--text pairs that belong to the `top' fashion category to match the VITON dataset characteristics.
The Fashion-Gen text descriptions were created by professional stylists and, as a result, contain many fashion-specific technical terms that are regarded difficult for general text--image matching models, such as CLIP due to the significant shift in comparison to the training data distribution. Thus, we only consider the sentences with a high match rate to the corresponding image. Specifically, we process each image and its corresponding description to obtain a CLIP matching score, then sort the image--text pairs by the match--score magnitude. We keep $120$ sentences with the highest matching scores, as these are the sentences that the CLIP model `understands' the best.
During testing, we combine each test image and with every text example to construct all possible image--text combinations, resulting in a total of $120 \cdot 120 = 14,400$ test combinations.

\renewcommand{\im}[1]{\includegraphics[width=\iwidth]{figures/dataset_examples/#1.jpg}}

\begin{figure}
    \centering
    \begin{tabular}{CCC}
    \im{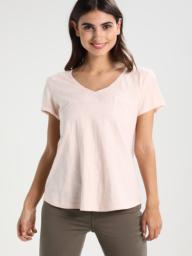} & \im{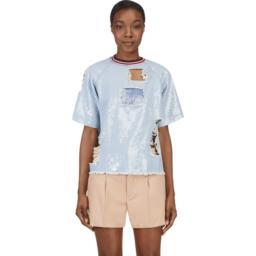} & \im{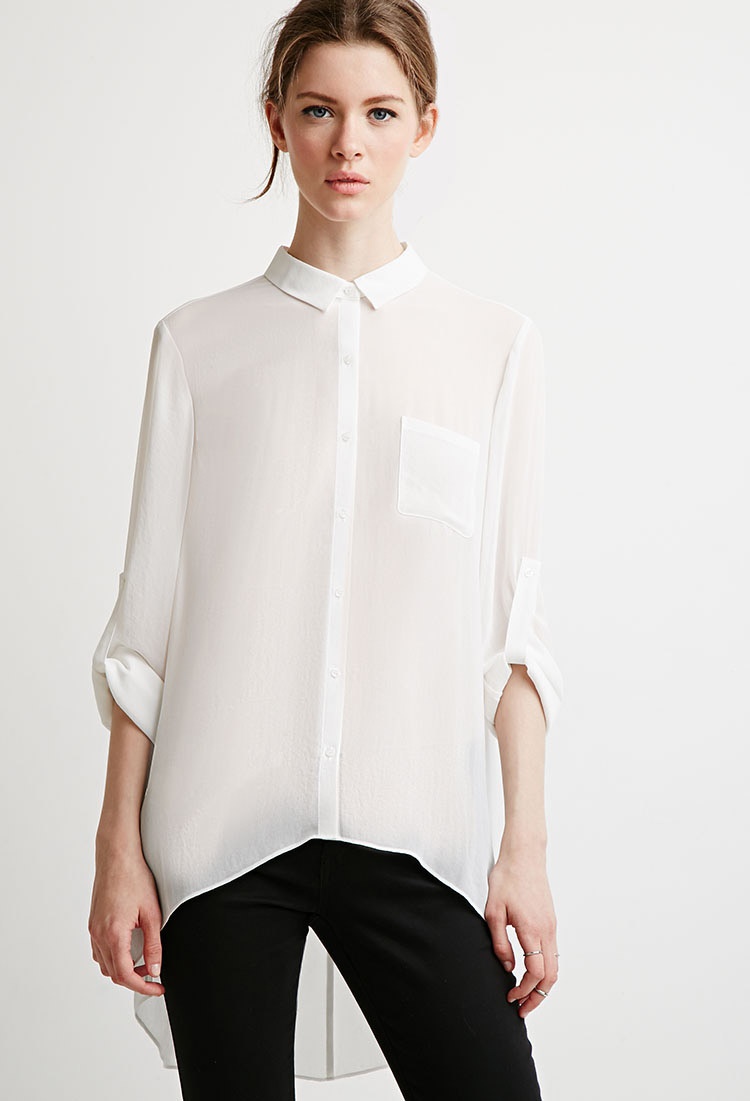} \\
    VITON & Fashion-Gen & DeepFashion \\
    \end{tabular}
    \caption{\textbf{Sample images from the experimental datasets.} We train and test FICE with fashion images from the VITON, Fashion-Gen and DeepFashion datasets.}
    \label{fig:dataset_examples}
\end{figure}


\subsection{Baseline Models}

We note that, to the best of our knowledge, no direct competitor addressing the same task has been presented in the literature so far. We, therefore, compare FICE to methods that fall under the umbrella of zero--shot text--based image editing~\cite{clip}. Specifically, we use various GAN inversion methods to obtain the latent code of the image, and then process the latent code using the global StyleCLIP  method~\cite{styleclip}, which modifies the latent code to enforce the desired semantics.

\textbf{GAN inversion.}
Given our trained GAN model, we train several GAN inversion encoder models to use as baselines in the experiments presented in the next section. Details on considered encoding techniques are given below:

\begin{itemize}
	\item \textbf{pSp}~\cite{psp}. The pSp model has been proposed for the tasks of conditional image synthesis, face frontalization, inpainting, and super-resolution and due to this broad application range has also been selected for our experiments. The architecture of pSp is based on a ResNet--style feature pyramid~\cite{fpn} and multiple encoder networks. The encoders predict a particular StyleGAN latent code from each convolutional layer of the feature pyramid and, in this way, embed images into the StyleGAN latent space.
	\item \textbf{E4e}~\cite{e4e}. E4e follows the architecture of the pSp model, but uses a distinct training process to ensure that the predicted (extended) latent code approximates the code defined in the original StyleGAN latent space. This allows the model to perform more convincing image manipulations when using techniques that try to alter the given latent code to achieve semantically meaningful edits. 
	\item \textbf{ReStyle}~\cite{restyle}. While pSp and E4e embed images in the latent space in a single (forward) pass, ReStyle uses an iterative procedure that gradually improves the embedding so it corresponds better to the given input image. We consider two ReStyle versions for the experiments, one with the pSp and one with the E4e encoder. 
	\item \textbf{HyperStyle}~\cite{hyperstyle}. Different from the inversion methods above, HyperStyle predicts the latent code of an input image, while also modifying the weights of the StyleGAN generator on a per-sample basis. The model first predicts an approximate latent code using the vanilla StyleGAN generator. This initial prediction, along with the original image, then serve as an input to a hyper-network that predicts offsets for the StyleGAN weights. The weights are finally modulated with the offsets and the resulting StyleGAN is used to synthesize the final image.
\end{itemize}

\renewcommand{\iwidth}{.135\textwidth}
\renewcommand{\im}[1]{\includegraphics[width=\iwidth]{figures/FICE_general_examples/#1.jpg}}
\renewcommand{\imrow}[1]{\im{#1} & \im{3/#1} & \im{2/#1} & \im{4/#1} & \im{0/#1} & \im{1/#1}  &  \im{6/#1}}


\begin{figure*}[t]
	\centering
	\begin{tabular}{CCCCCCC}
\imrow{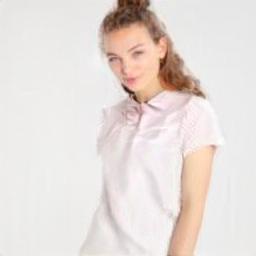} \\
\imrow{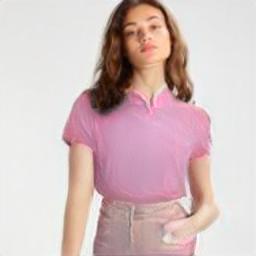} \\
\imrow{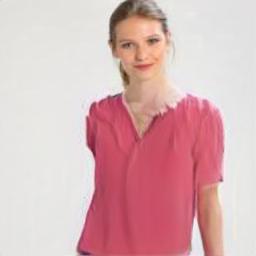} \\
\imrow{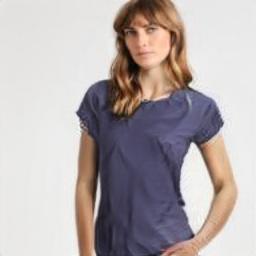} \\
Input & \scriptsize \textit{``Sleeveless hand-embroidered sequin camisole featuring a camo pattern throughout in tones of green, brown, and black''}	& \scriptsize \textit{``Long sleeve silk crepe de chine shirt featuring graphic pattern printed in tones of blue''} & \scriptsize \textit{``Short sleeve fine mercerized cotton piqu\'e polo in navy''} &   \scriptsize \textit{``Short sleeve ruffled cotton poplin and jersey blouse in pink''} & \scriptsize \textit{``Short sleeve cotton jersey t-shirt featuring floral pattern in tones of army green and off-white.''} & \scriptsize \textit{``Short sleeve cotton piqu\'e polo in soft black''} \\
	\end{tabular}
	\caption{\textbf{Example results generated by FICE for various text descriptions.} As can be seen, FICE is capable of synthesizing complex fashion styles (e.g., see the camouflage pattern in the  second column), while preserving the pose and identity of the subjects as well other image characteristics. Note the realism and seamless integration of different clothing items (in various designs, materials and shapes) without any 3D modelling. Best viewed electronically and zoomed-in.} 
	\label{fig:FICE_general_examples}
\end{figure*}

\textbf{StyleCLIP Implementation.}
To provide editing capabilities for the GAN inversion techniques, we use StyleCLIP's global editing method, which displaces the latent code of the given input image along a certain editing direction in accordance with the semantics encoded in some text description. 
StyleCLIP relies on two hyperparameters $(\alpha, \beta)$, where $\alpha$ defines the magnitude of the displacement in the latent space and $\beta$ represents a threshold that controls the entanglement of the edited attributes. 
The choice of $\alpha$ determines the strength of the  semantic content (from the text) present 
in the edited image. However, large values of $\alpha$ are known to degrade image quality and alter both the pose and identity of the input subject. These adverse effects can be reduced by using a disentanglement mechanism with a certain value of $\beta$. However, large  values of $\beta$ reduce the presence of the desired semantic presence. A suitable trade--off is therefore required.

We consider this trade--off in our comparative assessment and perform StyleCLIP experiments with several different combinations of $(\alpha, \beta)$ values. Specifically, we use $\alpha$ values of $\{3.0, 4.0, 5.0, 7.5, 10.0\}$ and $\beta$ values of $\{0.00, 0.025, 0.050, 0.075, 0.100, 0.125 \}$, which represent a reasonable cross-section of values for the evaluation. 

\subsection{Implementation Details}

The generator $G$ of FICE is based on StyleGAN2~\cite{stylegan2} and was trained on the training split of the VITON dataset. Prior to training, the image from the dataset were first cropped to $192 \times 192$ px by removing the bottom part of the image and then resized to $256 \times 256$ px using bilinear interpolation. The training was performed for a total of $450,000$ iterations, achieving a final Fr\'echet Inception Distance (FID) of $3.83$. 

For pose-parsing, we utilized the DensePose model~\cite{densepose} $D$ with a ResNet50 backbone and the Panoptic FPN head~\cite{kirillov2019panoptic}. We used the pre-trained model from the Detectron2 repository~\cite{detectron2}. The selected model is capable of parsing $24$ ind\-ividual body parts (e.g. upper left arm). For our implementation, we only considered body parts that are suitable (or applicable) for the VITON dataset, i.e., the upper body indices.

The segmentation model $S$ was trained on the DeepFashion dataset~\cite{deepfashion}. We used cross-entropy as the learning objective and weighted the predictions according to the class imbalance of the training split of the dataset. The learning procedure was performed for $19$ epochs (until convergence) using the Adam optimizer~\cite{adam} and a fixed learning rate of $\eta=10^{-4}$.

The GAN inversion E4e~\cite{e4e} encoder $E$ was trained on the VITON training dataset, using $G$ as the target GAN model to invert. Finally, 
the CLIP RN50x4 network architecture was chosen as the image encoder for the CLIP model. For the optimization--based GAN inversion procedure in Eq.~\eqref{eq: contrainedGANinv}, balancing weights were chosen  based on preliminary experiments and visual inspection of the results on the training data, so that $\lambda_{clip}=1, \lambda_{im}=30, \lambda_{pose}=10, \lambda_{head}=1, \lambda_{reg}=1$. The number of GAN inversion optimization iterations was fixed and set to $500$ for all experiments. We used the Adam optimization algorithm~\cite{adam}  with a learning rate of $\eta = 5 \cdot 10^{-2}$ when optimizing the latent code to compute $w^*$. 

\renewcommand{\twidth}{.12\textwidth}
\renewcommand{\iwidth}{.117\textwidth}
\renewcommand{\im}[1]{\includegraphics[width=\iwidth]{figures/qualitative-results/multi-sentence/#1.jpg}}
\renewcommand{\imrow}[1]{\im{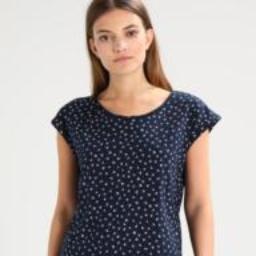} & \im{#1/psp} & \im{#1/e4e} & \im{#1/restyle-psp} & \im{#1/restyle-e4e} & \im{#1/hyperstyle} & \im{#1/fashprop}}

\begin{figure*}[t]
	\centering
	\begin{tabular}{DCCCCCCC}
\scriptsize \textit{``Short sleeve fine mercerized cotton piqué polo in navy.''} & \imrow{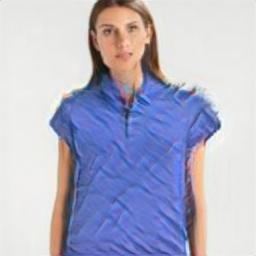} \\
\scriptsize \textit{``Long sleeve cotton sateen shirt in white.''}	& \imrow{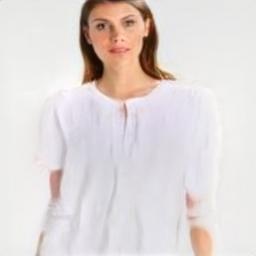} \\
\scriptsize \textit{``Short sleeve cotton piqué; polo in vivid red.''}	& \imrow{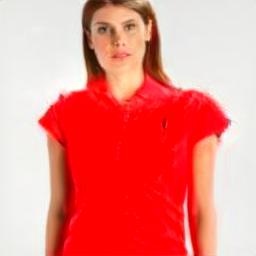} \\
\scriptsize \textit{``Sleeveless woven silk top in metallic silver-tone.''}	& \imrow{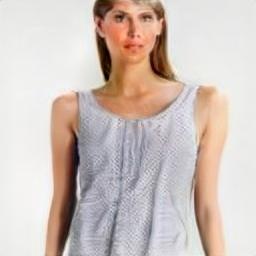} \\
Text description & Input image & pSp \cite{psp} & E4e \cite{e4e} & ReStyle-pSp \cite{restyle} & ReStyle-E4e \cite{restyle} & HyperStyle \cite{hyperstyle} & FICE (Ours) \\
	\end{tabular}
	\caption{\textbf{Example results with the \textit{Same image--different text} (SI-DT) configuration.} While most techniques are able to incorporate the desired semantics, FICE leads to the most realistic results and most convincingly preserves identity.}
	\label{fig:qualitative_results_1}
\end{figure*}


\section{Results and Discussion}

In this section, we report results that: $(i)$ demonstrate the capabilities of FICE for fashion image editing through several qualitative examples, $(ii)$ compare the proposed model to several competing techniques, $(iii)$ highlight the importance of various components through rigorous ablation studies, and $(iv)$ explore the model's limitations. Some additional results are available in the Supplementary material.

\subsection{Qualitative Results}

\textbf{FICE Evaluation.} We first evaluate the editing capabilities of FICE in Fig.~\ref{fig:FICE_general_examples} over a number of test images with different pose characteristics and  (initial) clothing styles (e.g., long and short sleeves, different material, designs, etc.) and different text descriptions. We observe that FICE is able to: $(i)$ synthesize complex clothing styles (see, for example, the camouflage pattern in the second column), $(ii)$ convincingly incorporate the semantics expressed in the text descriptions into the edited images, $(iii)$ preserve the pose and identity of the subjects, $(iv)$ add or remove sleeves from initial clothing (and hallucinate initially obscured objects, e.g., arms), and $(v)$ ensure a seamless fit with realistic clothing characteristics (e.g., with creases) without explicit 3D modelling.  The presented examples speak of the impressive editing capabilities of FICE and illustrate the flexibility of text--based editing. 

\renewcommand{\iwidth}{.135\textwidth}
\renewcommand{\im}[2]{\includegraphics[width=\iwidth]{figures/qualitative-results/single-sentence/#1/#2.jpg}}
\renewcommand{\imrow}[1]{\im{#1}{input} & \im{#1}{psp} & \im{#1}{e4e} & \im{#1}{restyle-psp} & \im{#1}{restyle-e4e} & \im{#1}{hyperstyle} & \im{#1}{fashprop}}
\begin{figure*}
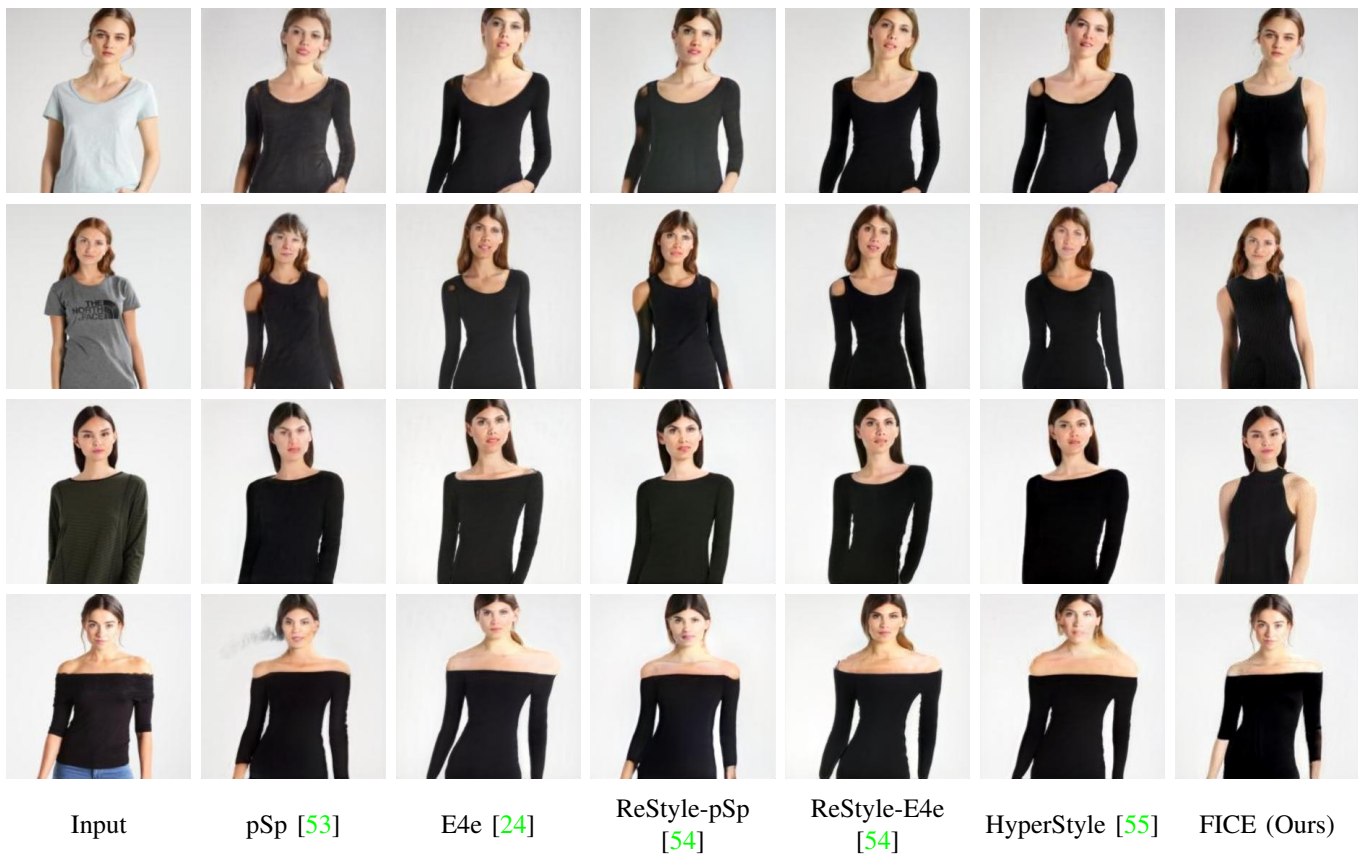
  
	\centering
    \begin{tabular}{CCCCCCC}
    \imrow{94} \\
    \imrow{62} \\
    \imrow{93} \\
    \imrow{51} \\
    Input & pSp \cite{psp} & E4e \cite{e4e} & ReStyle-pSp \cite{restyle} & ReStyle-E4e \cite{restyle} & HyperStyle \cite{hyperstyle} & FICE (Ours) \\
    \end{tabular}
	\caption{\textbf{Visual examples generated with the \textit{Same text--different image} (ST-DI) experimental configuration.} All models were tested with the same target description \textit{``Sleeveless rib knit wool bodysuit in black''}. Note how all models are able to infuse the targeted semantics, but except for FICE, often also produce pose/position changes as well as visual artifacts.}
	\label{fig:qualitative_results_2}
\end{figure*}

\textbf{Comparative Evaluation.} Next, we compare FICE to the competing models introduced in the previous section. For the evaluation, we consider two distinct cases:
\begin{itemize}
    \item \textit{Same image--different text (SI-DT).} In this configuration, we test all models on the same input image and pair it with different target text descriptions to demonstrate how the models handle different types of semantics.
    \item \textit{Same text--different image (ST-DI).} Here, we use different input images and pair them with the same target text description to explore the consistency of the edits made by the models with diverse inputs.
\end{itemize}

From the results in Fig.~\ref{fig:qualitative_results_1} (SI-DT) and Fig.~\ref{fig:qualitative_results_2} (ST-DI), we see that the competing models do not preserve identity nearly as well as FICE. The reason for that is twofold. First, when an image is embedded in the GAN latent space, there occurs some loss of identity information due to imperfect reconstruction through the generator $G$. The second reason lies in the entanglement of the latent code space that causes several undesired changes to the edited image. Even though StyleCLIP integrates a disentanglement mechanism, a shift of the latent code still partially affects the person's identity. FICE, on the other hand, preserves the identity due to the image stitching formulation of the model. Moreover, we observe that the editing results of the competing models exhibit certain entanglement issues, as can be seen well in the third row of Fig.~\ref{fig:qualitative_results_1}, where a change to red clothes also adds red lipstick. Finally, FICE performs better when editing certain clothing styles such as `polo shirt' as seen in a more convincing collar in first and third row of Fig.~\ref{fig:qualitative_results_1}. While the semantics from the target text can be seen in most edited images, the results for the proposed model are visually clearly the most convincing. 

When looking at Fig.~\ref{fig:qualitative_results_2}, we see that the competing models often have problems with exact positioning of the person in the edited image. The person is often shifted with respect to the original image, as seen best in the result of the E4e model in Fig. \ref{fig:qualitative_results_2}. Such pose shifts make it impossible to integrate image stitching techniques into the competing models. The main reason for the observed behavior can be attributed to the use of perceptual losses~\cite{johnson2016perceptual} when training the encoder models. The perceptual losses generally increase image fidelity at the expense of precise object localization. 
While such a trade--off is reasonable for other image processing tasks, it is not optimal for the task of Virtual Try-On, where the pose and position of the input person should be preserved exactly. FICE integrates a pose--preserving loss term with a pixel--level loss term to avoid such positioning problems. We note that all tested models generate consistent edits in terms of desired semantics, but except for FICE, also often introduce considerable visual artifacts.  



\subsection{Quantitative Results}
To be able to evaluate the performance of the generated images quantitatively, we define four distinct criteria that address different aspects of the editing process, i.e.:

\begin{figure*}[ht]
\includegraphics[width=\textwidth]{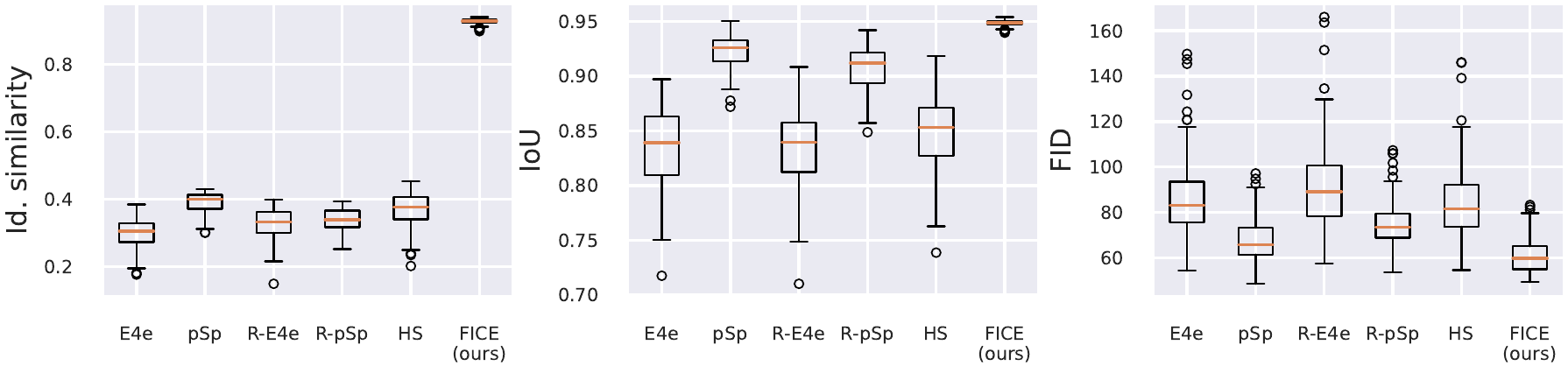}
    \caption{\textbf{Boxplots of different performance indicators for the tested editing models.} The competing models were tested with hyperparameter settings that resulted in the highest semantic--relevance score to ensure a fair comparison. Results are reported in terms of variation over the text descriptions. The individual scores are obtained by averaging over all images for a given text description. We observe that FICE performs best across all performance indices, while ensuring the most consistent results.} 
	\label{fig:whiskers}
\end{figure*}
\setlength{\tabcolsep}{2pt}
\renewcommand{\arraystretch}{1.2}

\begin{figure*}[ht]
\includegraphics[width=\textwidth]{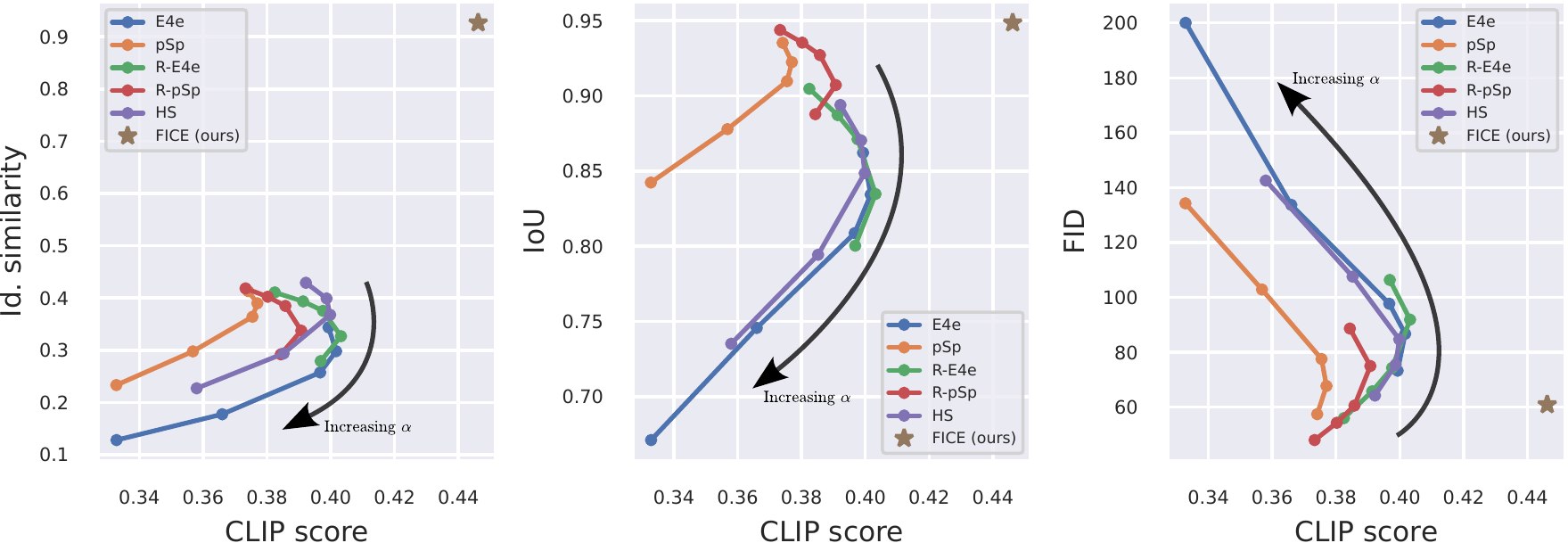}
    \caption{\textbf{Fine-grained comparison across multiple hyperparameter values.} For the StyleCLIP based models, we consider the $\beta$ value of the best-performing $(\alpha, \beta)$ combination and $\alpha \in \{3.0, 4.0, 5.0, 7.5, 10.0 \}$. Increasing $\alpha$  generally increases the semantic--relevance score, but above a certain point degrades the results. 
    }
	\label{fig:quantitative_results_graphics}
\end{figure*}

\begin{itemize}
    \item \textbf{Semantic Relevance.} The synthesized images should contain semantics that are relevant with respect to the input text $t$. We evaluate this aspect in the experiments with the CLIP model, defining a CLIP similarity score between a given image $I$ and the target text $t$ as, 
\begin{equation}
	S(I,t) = \cos (C^i(I), C^t(t)),
\end{equation}
where $C^i$ and $C^t$ are again the CLIP image and text encoders, respectively.
\item  \textbf{Pose Preservation.}
The edited image should preserve the pose of the input person. We evaluate pose--preservation capabilities using the Intersection over Union (IoU), a commonly used metric in the field of semantic segmentation~\cite{pascal_challenge}. 
We utilize DensePose for pose prediction because the model is insensitive with respect to overlaid clothing and compute an IoU score between the pose predictions from the input and edited images. 
\item \textbf{Identity Similarity.} The editing model should preserve the identity and facial appearance of the input image. We therefore use
the RetinaFace model~\cite{retinaface} to detect the face region, and then process it with ArcFace~\cite{arcface}, a face recognition model, to extract a face embedding vector. To quantify the similarity between the input and edited image in terms of identity, we compute the cosine similarity between the corresponding face embedding vectors.
\item \textbf{Image Fidelity.}
The editing methods should synthesize high fidelity images that are comparable to the inputs. To measure the quality of the generated images, we use the Fr\'echet Inception Distance (FID)~\cite{fid}. The metric is based on the difference between the statistics of image embeddings as extracted by the InceptionV3 neural network~\cite{szegedy2016rethinking} and has been used as one of the main metrics to evaluate the image quality of different image synthesis models~\cite{progan, stylegan1, stylegan2, stylegan2-ada}.
\end{itemize}

When calculating the final scores for performance reporting, the scores for images and text descriptions are averaged for the semantic relevance score, IoU, and identity similarity metrics. The FID score is averaged only over the text descriptions, since the metric is already calculated over a set of images.

In general, all methods exhibit some trade--off, where higher semantic relevance of the image is associated with worse performance on all other performance indicators considered. Therefore, we decide to evaluate the results for the competing methods for several sets of hyperparameter values. We directly compare FICE with the competitors at the hyperparameters values $(\alpha, \beta)$, where the maximum semantic score was obtained to ensure a reasonable and fair evaluation. 

The results of the outlined experiments are shown in Fig.~\ref{fig:whiskers} in terms of variability with respect to text descriptions and in Table~\ref{tab:quantitative_results} as average (overall) results. As can be seen, FICE significantly outperforms the competing methods in terms of all evaluated metrics. The image stitching technique helps FICE achieve a superior result in terms of identity preservation when compared to other models. Only the minute artifacts of the image stitching prevent FICE from achieving a perfect identity similarity score. In Fig.~\ref{fig:quantitative_results_graphics} we show the results for several $\alpha$ values, where the $\beta$ value is fixed to the value from the best $(\alpha, \beta)$ combination. Increasing the $\alpha$ value leads to a higher CLIP score (higher semantic relevance), but only up to a certain point, after which the results become worse. All other metrics worsen when the $\alpha$ value is increased.




\subsection{Ablation Study}
We ablate various components of FICE to study the impact of: $(i)$ the latent space choice made, $(ii)$ the initialization scheme used, and $(iii)$ the contribution of different loss terms. The motivation for the ablation is to further justify the design choices made with the proposed editing approach. 

\textbf{Latent Space and Initialization.}
To demonstrate the importance  of the latent space choice made for FICE as well as the importance of the  initialization procedure, we consider the vanilla $\mathcal{W}$ and the extended $\mathcal{W}^+$ latent code spaces, and initializations with either the latent code mean $w \leftarrow \bar{w}$ or the GAN inversion encoder predicted latent code $w \leftarrow E(I)$. Because the utilized GAN inversion $E$ encoder (E4e) is based on the extended $\mathcal{W}^+$ space, we train a separate E4e encoder (denoted as $\hat{E}$) for the vanilla latent code space $\mathcal{W}$ using the proposed method from the E4e repository\footnote{Available at \url{https://github.com/omertov/encoder4editing}}.


\renewcommand{\iwidth}{.30\columnwidth}
\renewcommand{\im}[1]{\includegraphics[width=\iwidth]{figures/ablation/init/#1.jpg}}
\renewcommand{\imrow}[1]{\im{input-#1} & \im{#1-vanilla} & \im{#1-wplus}}

\begin{figure}
    \centering
    \begin{tabular}{CCC}
        \imrow{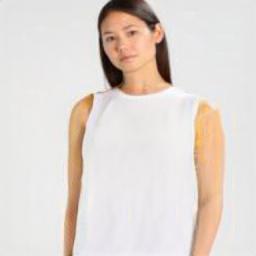} \\
        \imrow{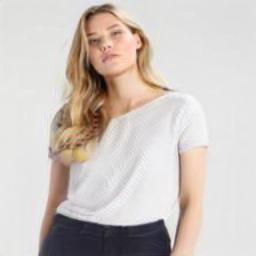} \\
        Input & $w \in \mathcal{W}$ & $w \in \mathcal{W^+}$ (FICE) \\
    \end{tabular}
    \caption{\textbf{Visual ablation study results -- latent space.} Examples of FICE--edited images when the latent space is restricted to $w \in \mathcal{W}$. The target description is set to \textit{``Structured knit bandeau top in yellow and off-white stripes''}. Due to lower expressiveness of the vanilla StyleGAN latent space $\mathcal{W}$, the  generated results exhibit limited pose preservation and semantic correctness.}
    \label{fig:ablation-init}
\end{figure}

From the results in Table~\ref{tab:ablation-init}, we see that the vanilla latent space $\mathcal{W}$ has limited expressive power, which adversely affects the semantic--relevance scores. The extended latent space $\mathcal{W}^+$ contributes towards higher semantic scores, but requires a careful initialization scheme. The trivial mean initialization scheme leads to a degradation in image fidelity, whereas a GAN--inversion based initialization leads to significant fidelity gains.  
We show visual examples that illustrate the impact of the latent--space choice 
in Fig.~\ref{fig:ablation-init}. Note how due to the limited expressiveness of the $\mathcal{W}$ space (compared to $\mathcal{W^+}$) the editing procedure is not able to infuse proper semantics into the images shown in the middle column. 

\renewcommand{\iwidth}{.18\columnwidth}
\renewcommand{\im}[2]{\includegraphics[width=\iwidth]{figures/ablation/loss-terms/#1/#2.jpg}}
\renewcommand{\imrow}[1]{\im{#1}{0} & \im{#1}{1} & \im{#1}{2} & \im{#1}{3} & \im{#1}{4}}
\setlength{\tabcolsep}{2pt}

\begin{figure}[t]
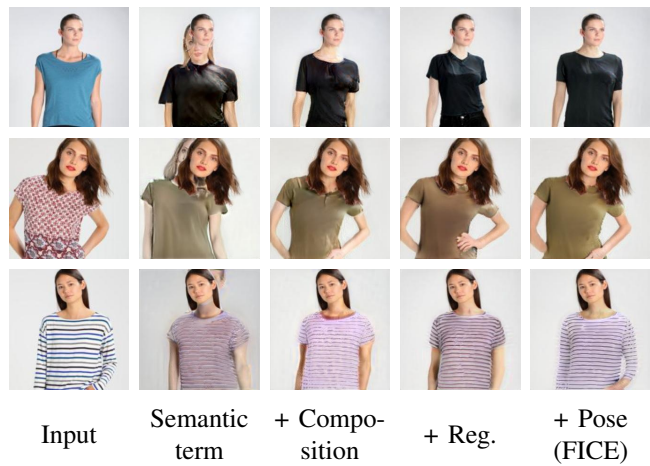

    \centering
    \begin{tabular}{CCCCC}
        \imrow{2} \\
        \imrow{3} \\
        \imrow{5} \\
        Input & Semantic term & + Composition & + Reg. & + Pose (FICE) \\
    \end{tabular}
    \caption{\textbf{Visual ablation study results -- loss terms.} An optimization objective that relies exclusively on the semantic term (2nd column) produces unsatisfactory image composition results. Adding the image composition term (3rd column) ensures pleasing image stitching results, but the output image still often contains visual artifacts. The addition of the latent--code regularization term (3rd column) helps mitigate these artifacts. Finally, the pose preservation term ensures more realistic and true results compared to the input.}
    \label{fig:ablation-loss_terms}
\end{figure}

\begin{table}[t]
	\centering
	\caption{\textbf{Quantitative comparison.} FICE is compared to several competing techniques and across four difference performance indicators. The arrows denote whether higher or lower scores imply better performance.} 
	\resizebox{0.999\columnwidth}{!}{%
	\begin{tabular}{l|cccccccr}
        \hline
        \hline
         \textbf{Model}                            && \textbf{Semantics} ($\uparrow$) && \textbf{Identity sim.} ($\uparrow$) && \textbf{IoU} ($\uparrow$)   && \textbf{FID} ($\downarrow$)     \\
        \hline
        pSp \cite{psp}               && $0.377$              && $0.390$                  && $0.922$          && $67.74$          \\
        e4e \cite{e4e}               && $0.402$              && $0.298$                  && $0.834$          && $86.74$          \\
        ReStyle-pSp \cite{restyle}   && $0.403$              && $0.337$                  && $0.907$          && $75.01$          \\
        ReStyle-e4e \cite{restyle}   && $0.391$              && $0.327$                  && $0.835$          && $91.92$          \\
        HyperStyle \cite{hyperstyle} && $0.399$              && $0.368$                  && $0.849$          && $84.70$          \\
        FICE (Ours)                  && $\mathbf{0.446}$     && $\mathbf{0.926}$         && $\mathbf{0.949}$ && $\mathbf{60.96}$ \\
        \hline
        \hline 
	\end{tabular}
	}
	\label{tab:quantitative_results}
\end{table}

\setlength{\tabcolsep}{2pt}
\begin{table}
	\centering
    \caption{\textbf{Ablation study w.r.t. latent spaces.} We analyze the choice of the latent space (vanilla $\mathcal{W}$ and extended $\mathcal{W}^+$) as well as the procedure for initializing the latent code $w$.}
    \resizebox{1\columnwidth}{!}{%
	\begin{tabular}{ll|cccccccrr|r}
        \hline \hline
             \textbf{Latent space variant}                             &&&  \textbf{Semantics} ($\uparrow$) && \textbf{Id. sim.} ($\uparrow$) && \textbf{IoU} ($\uparrow$)     && \textbf{FID} ($\downarrow$) \\
                                          \hline
        $\bar{w}, \mathcal{W}$          &&& $0.422$             &&$0.923$             && $0.914$           && $72.11$         \\
        $\bar{w}, \mathcal{W}^+$        &&& $0.443$             &&$0.918$             && $0.919$           && $77.13$         \\
       $\hat{E}(I), \mathcal{W}$         &&& $0.378$             &&$0.911$             && $0.886$           && $63.16$         \\
       $E(I), \mathcal{W}^+$ (FICE) &&& $\mathbf{0.446}$    && $\mathbf{0.926}$   && $\mathbf{0.949}$  && $\mathbf{60.96}$\\
        \hline \hline
	\end{tabular}
	}
	\label{tab:ablation-init}
\end{table}

\textbf{Loss Function.}
Next, we evaluate the contribution of the individual loss terms from Eq.~\eqref{eq:final} on the generated results. Specifically, we ablate three distinct terms  by setting the corresponding loss weights to $0$: $(i)$ the pose preservation term ($\lambda_{pose}=0$), $(ii)$ the image composition term ($\lambda_{im}=0$ and $\lambda_{head} = 0$), and $(iii)$ the latent--code regularization term ($\lambda_{reg}=0$). We keep the semantic-related CLIP term as text--based editing is not possible without it.

\setlength{\tabcolsep}{2.5pt}
\renewcommand{\arraystretch}{1.2}
\begin{table}
	\centering
    \caption{\textbf{Ablation study results w.r.t. loss terms.} We analyze the impact of individual FICE optimization objectives across four performance measures.} 
    \resizebox{1\columnwidth}{!}{%
	\begin{tabular}{lc|cccccccr}
        \hline \hline
       \textbf{Objective} &&&  \textbf{Semantics} ($\uparrow$) && \textbf{Id. sim.} ($\uparrow$) && \textbf{IoU} ($\uparrow$) && \textbf{FID} ($\downarrow$) \\
		\hline
        Semantic term ($\mathcal{L}_{clip}$) &&& $0.453$ && $0.819$ && $0.781$ && $82.52$ \\
        + composition ($\mathcal{L}_{im/head}$) &&& $\mathbf{0.466}$ && $0.924$ && $0.888$ && $68.94$ \\ 
        + latent reg. ($\mathcal{L}_{reg}$) &&& $0.450$ && $0.924$ && $0.884$ && $64.14$ \\
        + pose pres. ($\mathcal{L}_{pose}$)$^{\dagger}$  &&& $0.446$ && $\mathbf{0.926}$ && $\mathbf{0.949}$ && $\mathbf{60.96}$ \\
        \hline \hline
        $^{\dagger}$ Complete FICE
	\end{tabular}
	\label{tab:ablation-loss}
	}
\end{table}

The results of loss-related ablations are shown quantitatively in Table~\ref{tab:ablation-loss} and qualitatively in Fig.~\ref{fig:ablation-loss_terms}. We observe that the image composition term significantly improves all metrics by ensuring that composition does not cause unnatural visual artifacts. Interestingly, the absence of the composition term strongly degrades the identity similarity score, even though the face is preserved by the image--stitching step. Regularization of the latent code slightly degrades the semantic--relevance score, but contributes to image fidelity. Finally, the pose preservation term also slightly lowers the semantic--relevance score, but contributes significantly to the IoU index and further improves the FID score.

\renewcommand{\iwidth}{.32\columnwidth}
\renewcommand{\im}[2]{\includegraphics[width=\iwidth]{figures/limitations/#1/#2.jpg}}

\subsection{Limitations} 

\begin{figure}[t]
    \begin{tabular}{CCC}
    \im{0}{084} & \im{1}{084} & \im{2}{084} \\
    \im{0}{106} & \im{1}{106} & \im{2}{106} \\
    \im{0}{115} & \im{1}{115} & \im{2}{115} \\
\scriptsize \textit{``Skull graphic at front in black''} & \scriptsize \textit{``Multi-colour lightning bolt print at front''} & \scriptsize \textit{``Saints Pauls cathedral landmark graphic printed in black and grey at front.''} \\
    \end{tabular}
	\caption{\textbf{Illustration of FICE limitations.} When presented with text descriptions involving objects and specific logos to be shown  on the garment, the model is only capable of generating approximate results. In the presented examples, the model fails to produce a convincing skull graphic (left column), a lightning bolt (middle column) or a cathedral (right column).}
	\label{fig:limitations}
\end{figure}

While the proposed FICE model achieves high-quality competitive image editing results, it still exhibits certain limitations. Specifically, the model inherits the constraints of CLIP on the length of the text description, which is limited to $76$ tokens, as extracted by the byte-pair encoding technique~\cite{bpe}. Furthermore, FICE uses gradient optimization as the basis for processing the fashion images. We batch the images to process 20 images at a time for a given text description, which takes about 5 minutes or 16 seconds per image on average (measured using NVIDIA GeForce RTX 3090). Meanwhile, an encoder--based method typically retrieves a latent code in less than a second and when followed by StyleCLIP's global direction method generates a new latent vector in less than an additional $100$ ms. Finally, with respect to the text description semantics, FICE sometimes struggles with sleeve length (see 3rd row of Fig.~\ref{fig:FICE_general_examples}) as well as text descriptions that contain detailed descriptions of objects/logos to be shown on the garment, as exemplified in Fig.~\ref{fig:limitations}. We suspect that the reason for the limited performance in such cases is a combination of limited GAN synthesis capabilities and limited CLIP knowledge. A possible solution to this problem could be to extend the GAN training dataset and fine-tune CLIP on large text--image fashion datasets.

\section{Conclusion}
In this paper, we presented FICE, a novel model for realistic text--conditioned fashion image editing. The core of the approach is based on GAN latent code optimization guided by CLIP semantic knowledge as well as pose, regularization, and composition constraints. We showed through rigorous experiments that FICE is able to convincingly edit fashion images, and that it significantly outperforms competing methods on all evaluated metrics, especially when it comes to identity preservation and semantic relevance of the edited images.

\section*{Acknowledgements}

Supported in parts by the Slovenian Research Agency ARRS through the Research Programme P2-0250(B) Metrology and Biometric System, the ARRS Project J2-2501(A) DeepBeauty and the ARRS junior researcher program.


\bibliographystyle{IEEEtran}
\bibliography{bibliography}  

\begin{thebibliography}{10}
\providecommand{\url}[1]{#1}
\csname url@samestyle\endcsname
\providecommand{\newblock}{\relax}
\providecommand{\bibinfo}[2]{#2}
\providecommand{\BIBentrySTDinterwordspacing}{\spaceskip=0pt\relax}
\providecommand{\BIBentryALTinterwordstretchfactor}{4}
\providecommand{\BIBentryALTinterwordspacing}{\spaceskip=\fontdimen2\font plus
\BIBentryALTinterwordstretchfactor\fontdimen3\font minus
  \fontdimen4\font\relax}
\providecommand{\BIBforeignlanguage}[2]{{%
\expandafter\ifx\csname l@#1\endcsname\relax
\typeout{** WARNING: IEEEtran.bst: No hyphenation pattern has been}%
\typeout{** loaded for the language `#1'. Using the pattern for}%
\typeout{** the default language instead.}%
\else
\language=\csname l@#1\endcsname
\fi
#2}}
\providecommand{\BIBdecl}{\relax}
\BIBdecl

\bibitem{fashion_industry}
A.~Kozlowski, M.~Bardecki, and C.~Searcy, ``Environmental impacts in the
  fashion industry: A life-cycle and stakeholder framework,'' \emph{Journal of
  Corporate Citizenship}, no.~45, pp. 17--36, 2012.

\bibitem{viton}
X.~Han, Z.~Wu, Z.~Wu, R.~Yu, and L.~S. Davis, ``Viton: An image-based virtual
  try-on network,'' in \emph{Computer Vision and Pattern Recognition (CVPR)},
  2018, pp. 7543--7552.

\bibitem{cp-vton}
B.~Wang, H.~Zheng, X.~Liang, Y.~Chen, L.~Lin, and M.~Yang, ``Toward
  characteristic-preserving image-based virtual try-on network,'' in
  \emph{European Conference on Computer Vision (ECCV)}, 2018, pp. 589--604.

\bibitem{vtnfp}
R.~Yu, X.~Wang, and X.~Xie, ``Vtnfp: An image-based virtual try-on network with
  body and clothing feature preservation,'' in \emph{International Conference
  on Computer Vision (ICCV)}, 2019, pp. 10\,511--10\,520.

\bibitem{yang2020towards}
H.~Yang, R.~Zhang, X.~Guo, W.~Liu, W.~Zuo, and P.~Luo, ``Towards
  photo-realistic virtual try-on by adaptively generating-preserving image
  content,'' in \emph{Computer Vision and Pattern Recognition (CVPR)}, 2020,
  pp. 7850--7859.

\bibitem{fele2022c}
B.~Fele, A.~Lampe, P.~Peer, and V.~Struc, ``C-vton: Context-driven image-based
  virtual try-on network,'' in \emph{IEEE/CVF Winter Conference on Applications
  of Computer Vision}, 2022, pp. 3144--3153.

\bibitem{mg-vton}
H.~Dong, X.~Liang, X.~Shen, B.~Wang, H.~Lai, J.~Zhu, Z.~Hu, and J.~Yin,
  ``Towards multi-pose guided virtual try-on network,'' in \emph{International
  Conference on Computer Vision (ICCV)}, 2019, pp. 9026--9035.

\bibitem{cyclefashion}
C.~Ge, Y.~Song, Y.~Ge, H.~Yang, W.~Liu, and P.~Luo, ``Disentangled cycle
  consistency for highly-realistic virtual try-on,'' in \emph{Computer Vision
  and Pattern Recognition (CVPR)}, 2021, pp. 16\,928--16\,937.

\bibitem{wuton}
T.~Issenhuth, J.~Mary, and C.~Calauz{\`e}nes, ``Do not mask what you do not
  need to mask: a parser-free virtual try-on,'' in \emph{European Conference on
  Computer Vision}.\hskip 1em plus 0.5em minus 0.4em\relax Springer, 2020, pp.
  619--635.

\bibitem{pf-afn}
Y.~Ge, Y.~Song, R.~Zhang, C.~Ge, W.~Liu, and P.~Luo, ``Parser-free virtual
  try-on via distilling appearance flows,'' in \emph{Computer Vision and
  Pattern Recognition (CVPR)}, 2021, pp. 8485--8493.

\bibitem{gan}
I.~Goodfellow, J.~Pouget-Abadie, M.~Mirza, B.~Xu, D.~Warde-Farley, S.~Ozair,
  A.~Courville, and Y.~Bengio, ``Generative adversarial nets,'' in
  \emph{Advances in neural information processing systems (NIPS)}, 2014.

\bibitem{prada}
S.~Zhu, R.~Urtasun, S.~Fidler, D.~Lin, and C.~Change~Loy, ``Be your own prada:
  Fashion synthesis with structural coherence,'' in \emph{International
  Conference on Computer Vision (ICCV)}, 2017, pp. 1680--1688.

\bibitem{text2human}
Y.~Jiang, S.~Yang, H.~Qiu, W.~Wu, C.~C. Loy, and Z.~Liu, ``Text2human:
  Text-driven controllable human image generation,'' \emph{arXiv preprint
  arXiv:2205.15996}, 2022.

\bibitem{clip}
A.~Radford, J.~W. Kim, C.~Hallacy, A.~Ramesh, G.~Goh, S.~Agarwal, G.~Sastry,
  A.~Askell, P.~Mishkin, J.~Clark \emph{et~al.}, ``Learning transferable visual
  models from natural language supervision,'' \emph{arXiv preprint
  arXiv:2103.00020}, 2021.

\bibitem{slip}
N.~Mu, A.~Kirillov, D.~Wagner, and S.~Xie, ``Slip: Self-supervision meets
  language-image pre-training,'' in \emph{European Conference on Computer
  Vision (ECCV)}.\hskip 1em plus 0.5em minus 0.4em\relax Springer, 2022, pp.
  529--544.

\bibitem{dalle}
A.~Ramesh, M.~Pavlov, G.~Goh, S.~Gray, C.~Voss, A.~Radford, M.~Chen, and
  I.~Sutskever, ``Zero-shot text-to-image generation,'' in \emph{International
  Conference on Machine Learning}.\hskip 1em plus 0.5em minus 0.4em\relax PMLR,
  2021, pp. 8821--8831.

\bibitem{dalle2}
A.~Ramesh, P.~Dhariwal, A.~Nichol, C.~Chu, and M.~Chen, ``Hierarchical
  text-conditional image generation with clip latents,'' \emph{arXiv preprint
  arXiv:2204.06125}, 2022.

\bibitem{ju2022prompting}
C.~Ju, T.~Han, K.~Zheng, Y.~Zhang, and W.~Xie, ``Prompting visual-language
  models for efficient video understanding,'' in \emph{European Conference on
  Computer Vision (ECCV)}.\hskip 1em plus 0.5em minus 0.4em\relax Springer,
  2022, pp. 105--124.

\bibitem{styleclip}
O.~Patashnik, Z.~Wu, E.~Shechtman, D.~Cohen-Or, and D.~Lischinski, ``Styleclip:
  Text-driven manipulation of stylegan imagery,'' in \emph{International
  Conference on Computer Vision (ICCV)}, 2021, pp. 2085--2094.

\bibitem{gan_inversion_survey}
W.~Xia, Y.~Zhang, Y.~Yang, J.-H. Xue, B.~Zhou, and M.-H. Yang, ``{GAN
  inversion: A Survey},'' \emph{IEEE Transactions on Pattern Analysis and
  Machine Intelligence}, 2022.

\bibitem{interfacegan}
Y.~Shen, C.~Yang, X.~Tang, and B.~Zhou, ``Interfacegan: Interpreting the
  disentangled face representation learned by gans,'' \emph{IEEE Transactions
  on Pattern Analysis and Machine Intelligence}, 2021.

\bibitem{abdal2021styleflow}
R.~Abdal, P.~Zhu, N.~J. Mitra, and P.~Wonka, ``Styleflow: Attribute-conditioned
  exploration of stylegan-generated images using conditional continuous
  normalizing flows,'' \emph{ACM Transactions on Graphics (ToG)}, vol.~40,
  no.~3, pp. 1--21, 2021.

\bibitem{harkonen2020ganspace}
E.~H{\"a}rk{\"o}nen, A.~Hertzmann, J.~Lehtinen, and S.~Paris, ``Ganspace:
  Discovering interpretable gan controls,'' \emph{Advances in Neural
  Information Processing Systems}, vol.~33, pp. 9841--9850, 2020.

\bibitem{e4e}
O.~Tov, Y.~Alaluf, Y.~Nitzan, O.~Patashnik, and D.~Cohen-Or, ``Designing an
  encoder for stylegan image manipulation,'' \emph{ACM Transactions on Graphics
  (TOG)}, vol.~40, no.~4, pp. 1--14, 2021.

\bibitem{s-space}
Z.~Wu, D.~Lischinski, and E.~Shechtman, ``Stylespace analysis: Disentangled
  controls for stylegan image generation,'' in \emph{Computer Vision and
  Pattern Recognition (CVPR)}, 2021, pp. 12\,863--12\,872.

\bibitem{maskfacegan}
M.~Pernu{\v{s}}, V.~{\v{S}}truc, and S.~Dobri{\v{s}}ek, ``High resolution face
  editing with masked gan latent code optimization,'' \emph{arXiv:2103.11135},
  2021.

\bibitem{fashiongen}
N.~Rostamzadeh, S.~Hosseini, T.~Boquet, W.~Stokowiec, Y.~Zhang, C.~Jauvin, and
  C.~Pal, ``Fashion-gen: The generative fashion dataset and challenge,''
  \emph{arXiv preprint arXiv:1806.08317}, 2018.

\bibitem{wu2017survey}
X.~Wu, K.~Xu, and P.~Hall, ``A survey of image synthesis and editing with
  generative adversarial networks,'' \emph{Tsinghua Science and Technology},
  vol.~22, no.~6, pp. 660--674, 2017.

\bibitem{cheng2021fashion}
W.-H. Cheng, S.~Song, C.-Y. Chen, S.~C. Hidayati, and J.~Liu, ``Fashion meets
  computer vision: A survey,'' \emph{ACM Computing Surveys (CSUR)}, vol.~54,
  no.~4, pp. 1--41, 2021.

\bibitem{dcgan}
A.~Radford, L.~Metz, and S.~Chintala, ``Unsupervised representation learning
  with deep convolutional generative adversarial networks,'' in
  \emph{International Conference on Learning Representations (ICLR)}, 2016.

\bibitem{progan}
T.~Karras, T.~Aila, S.~Laine, and J.~Lehtinen, ``Progressive growing of gans
  for improved quality, stability, and variation,'' in \emph{International
  Conference on Learning Representations (ICLR)}, 2018.

\bibitem{stylegan1}
T.~Karras, S.~Laine, and T.~Aila, ``A style-based generator architecture for
  generative adversarial networks,'' in \emph{Computer Vision and Pattern
  Recognition (CVPR)}, 2019, pp. 4401--4410.

\bibitem{stylegan2}
T.~Karras, S.~Laine, M.~Aittala, J.~Hellsten, J.~Lehtinen, and T.~Aila,
  ``Analyzing and improving the image quality of stylegan,'' in \emph{Computer
  Vision and Pattern Recognition (CVPR)}, 2020, pp. 8110--8119.

\bibitem{stylegan2-ada}
T.~Karras, M.~Aittala, J.~Hellsten, S.~Laine, J.~Lehtinen, and T.~Aila,
  ``Training generative adversarial networks with limited data,''
  \emph{Advances in Neural Information Processing Systems}, pp.
  12\,104--12\,114, 2020.

\bibitem{stylegan3}
T.~Karras, M.~Aittala, S.~Laine, E.~H{\"a}rk{\"o}nen, J.~Hellsten, J.~Lehtinen,
  and T.~Aila, ``Alias-free generative adversarial networks,'' \emph{Advances
  in Neural Information Processing Systems}, vol.~34, pp. 852--863, 2021.

\bibitem{wgan}
M.~Arjovsky, S.~Chintala, and L.~Bottou, ``Wasserstein generative adversarial
  networks,'' in \emph{International Conference on Machine Learning (ICML)},
  2017, pp. 214--223.

\bibitem{wgan-gp}
I.~Gulrajani, F.~Ahmed, M.~Arjovsky, V.~Dumoulin, and A.~C. Courville,
  ``Improved training of wasserstein gans,'' in \emph{Advances in Neural
  Information Processing Systems (NIPS)}, 2017, pp. 5767--5777.

\bibitem{lsgan}
X.~Mao, Q.~Li, H.~Xie, R.~Y. Lau, Z.~Wang, and S.~Paul~Smolley, ``Least squares
  generative adversarial networks,'' in \emph{International Conference on
  Computer Vision (ICCV)}, 2017, pp. 2794--2802.

\bibitem{miyato2018spectral}
T.~Miyato, T.~Kataoka, M.~Koyama, and Y.~Yoshida, ``Spectral normalization for
  generative adversarial networks,'' in \emph{International Conference on
  Learning Representations (ICLR)}, 2018.

\bibitem{mescheder2018training}
L.~Mescheder, A.~Geiger, and S.~Nowozin, ``Which training methods for gans do
  actually converge?'' in \emph{International Conference on Machine learning
  (ICML)}, 2018, pp. 3481--3490.

\bibitem{reed2016generative}
S.~Reed, Z.~Akata, X.~Yan, L.~Logeswaran, B.~Schiele, and H.~Lee, ``Generative
  adversarial text to image synthesis,'' in \emph{International conference on
  machine learning}.\hskip 1em plus 0.5em minus 0.4em\relax PMLR, 2016, pp.
  1060--1069.

\bibitem{stackgan}
H.~Zhang, T.~Xu, H.~Li, S.~Zhang, X.~Wang, X.~Huang, and D.~N. Metaxas,
  ``Stackgan: Text to photo-realistic image synthesis with stacked generative
  adversarial networks,'' in \emph{International Conference on Computer Vision
  (ICCV)}, 2017, pp. 5907--5915.

\bibitem{stackgan++}
------, ``Stackgan++: Realistic image synthesis with stacked generative
  adversarial networks,'' \emph{IEEE transactions on pattern analysis and
  machine intelligence}, vol.~41, no.~8, pp. 1947--1962, 2018.

\bibitem{attngan}
T.~Xu, P.~Zhang, Q.~Huang, H.~Zhang, Z.~Gan, X.~Huang, and X.~He, ``Attngan:
  Fine-grained text to image generation with attentional generative adversarial
  networks,'' in \emph{Computer Vision and Pattern Recognition (CVPR)}, 2018,
  pp. 1316--1324.

\bibitem{mirrorgan}
T.~Qiao, J.~Zhang, D.~Xu, and D.~Tao, ``Mirrorgan: Learning text-to-image
  generation by redescription,'' in \emph{Computer Vision and Pattern
  Recognition (CVPR)}, 2019, pp. 1505--1514.

\bibitem{imagen}
C.~Saharia, W.~Chan, S.~Saxena, L.~Li, J.~Whang, E.~Denton, S.~K.~S.
  Ghasemipour, B.~K. Ayan, S.~S. Mahdavi, R.~G. Lopes \emph{et~al.},
  ``Photorealistic text-to-image diffusion models with deep language
  understanding,'' \emph{arXiv preprint arXiv:2205.11487}, 2022.

\bibitem{dong2017semantic}
H.~Dong, S.~Yu, C.~Wu, and Y.~Guo, ``Semantic image synthesis via adversarial
  learning,'' in \emph{International Conference on Computer Vision (ICCV)},
  2017, pp. 5706--5714.

\bibitem{nam2018text}
S.~Nam, Y.~Kim, and S.~J. Kim, ``Text-adaptive generative adversarial networks:
  manipulating images with natural language,'' \emph{Advances in neural
  information processing systems}, vol.~31, 2018.

\bibitem{li2020manigan}
B.~Li, X.~Qi, T.~Lukasiewicz, and P.~H. Torr, ``Manigan: Text-guided image
  manipulation,'' in \emph{Computer Vision and Pattern Recognition (CVPR)},
  2020, pp. 7880--7889.

\bibitem{image2stylegan}
R.~Abdal, Y.~Qin, and P.~Wonka, ``Image2stylegan: How to embed images into the
  stylegan latent space?'' in \emph{International Conference on Computer Vision
  (ICCV)}, 2019, pp. 4431--4440.

\bibitem{image2stylegan++}
------, ``Image2stylegan++: How to edit the embedded images?'' in
  \emph{Computer Vision and Pattern Recognition (CVPR)}, 2020.

\bibitem{tedigan}
W.~Xia, Y.~Yang, J.-H. Xue, and B.~Wu, ``Tedigan: Text-guided diverse face
  image generation and manipulation,'' in \emph{Computer Vision and Pattern
  Recognition}, 2021, pp. 2256--2265.

\bibitem{psp}
E.~Richardson, Y.~Alaluf, O.~Patashnik, Y.~Nitzan, Y.~Azar, S.~Shapiro, and
  D.~Cohen-Or, ``Encoding in style: a stylegan encoder for image-to-image
  translation,'' in \emph{Computer Vision and Pattern Recognition (CVPR)},
  2021, pp. 2287--2296.

\bibitem{restyle}
Y.~Alaluf, O.~Patashnik, and D.~Cohen-Or, ``Restyle: A residual-based stylegan
  encoder via iterative refinement,'' in \emph{International Conference on
  Computer Vision (ICCV)}, 2021, pp. 6711--6720.

\bibitem{hyperstyle}
Y.~Alaluf, O.~Tov, R.~Mokady, R.~Gal, and A.~H. Bermano, ``Hyperstyle: Stylegan
  inversion with hypernetworks for real image editing,'' \emph{arXiv preprint
  arXiv:2111.15666}, 2021.

\bibitem{cycle}
J.-Y. Zhu, T.~Park, P.~Isola, and A.~A. Efros, ``Unpaired image-to-image
  translation using cycle-consistent adversarial networks,'' in
  \emph{International Conference on Computer Vision (ICCV)}, 2017, pp.
  2223--2232.

\bibitem{densepose}
R.~A. G{\"u}ler, N.~Neverova, and I.~Kokkinos, ``Densepose: Dense human pose
  estimation in the wild,'' in \emph{Computer Vision and Pattern Recognition
  (CVPR)}, 2018, pp. 7297--7306.

\bibitem{deeplabv3}
L.-C. Chen, G.~Papandreou, F.~Schroff, and H.~Adam, ``Rethinking atrous
  convolution for semantic image segmentation,'' in \emph{Computer Vision and
  Pattern Recognition (CVPR)}, 2017.

\bibitem{deepfashion}
Z.~Liu, P.~Luo, S.~Qiu, X.~Wang, and X.~Tang, ``Deepfashion: Powering robust
  clothes recognition and retrieval with rich annotations,'' in \emph{Computer
  Vision and Pattern Recognition (CVPR)}, 2016, pp. 1096--1104.

\bibitem{pulse}
S.~Menon, A.~Damian, S.~Hu, N.~Ravi, and C.~Rudin, ``Pulse: Self-supervised
  photo upsampling via latent space exploration of generative models,'' in
  \emph{Computer Vision and Pattern Recognition (CVPR)}, 2020, pp. 2437--2445.

\bibitem{fpn}
T.-Y. Lin, P.~Doll{\'a}r, R.~Girshick, K.~He, B.~Hariharan, and S.~Belongie,
  ``Feature pyramid networks for object detection,'' in \emph{Computer Vision
  and Pattern Recognition (CVPR)}, 2017, pp. 2117--2125.

\bibitem{kirillov2019panoptic}
A.~Kirillov, R.~Girshick, K.~He, and P.~Doll{\'a}r, ``Panoptic feature pyramid
  networks,'' in \emph{Computer Vision and Pattern Recognition (CVPR)}, 2019,
  pp. 6399--6408.

\bibitem{detectron2}
Y.~Wu, A.~Kirillov, F.~Massa, W.-Y. Lo, and R.~Girshick, ``Detectron2,''
  \url{https://github.com/facebookresearch/detectron2}, 2019.

\bibitem{adam}
D.~P. Kingma and J.~Ba, ``Adam: A method for stochastic optimization,'' in
  \emph{International Conference on Learning Representations (ICLR)}, 2014.

\bibitem{johnson2016perceptual}
J.~Johnson, A.~Alahi, and L.~Fei-Fei, ``Perceptual losses for real-time style
  transfer and super-resolution,'' in \emph{European conference on computer
  vision}, 2016, pp. 694--711.

\bibitem{pascal_challenge}
M.~Everingham, S.~M.~A. Eslami, L.~Van~Gool, C.~K. Williams, J.~Winn, and
  A.~Zisserman, ``\BIBforeignlanguage{en}{The pascal visual object classes
  challenge: A retrospective},'' \emph{\BIBforeignlanguage{en}{International
  Journal of Computer Vision}}, vol. 111, no.~1, pp. 98 -- 136, 2015.

\bibitem{retinaface}
J.~Deng, J.~Guo, E.~Ververas, I.~Kotsia, and S.~Zafeiriou, ``Retinaface:
  Single-shot multi-level face localisation in the wild,'' in \emph{Computer
  Vision and Pattern Recognition (CVPR)}, June 2020.

\bibitem{arcface}
J.~Deng, J.~Guo, N.~Xue, and S.~Zafeiriou, ``Arcface: Additive angular margin
  loss for deep face recognition,'' in \emph{Computer Vision and Pattern
  Recognition (CVPR)}, 2019, pp. 4690--4699.

\bibitem{fid}
M.~Heusel, H.~Ramsauer, T.~Unterthiner, B.~Nessler, and S.~Hochreiter, ``Gans
  trained by a two time-scale update rule converge to a local nash
  equilibrium,'' in \emph{Advances in neural information processing systems
  (NIPS)}, 2017, pp. 6626--6637.

\bibitem{szegedy2016rethinking}
C.~Szegedy, V.~Vanhoucke, S.~Ioffe, J.~Shlens, and Z.~Wojna, ``Rethinking the
  inception architecture for computer vision,'' in \emph{Computer Vision and
  Pattern Recognition (CVPR)}, 2016, pp. 2818--2826.

\bibitem{bpe}
R.~Sennrich, B.~Haddow, and A.~Birch, ``Neural machine translation of rare
  words with subword units,'' \emph{arXiv preprint arXiv:1508.07909}, 2015.

\bibitem{dong2019towards}
H.~Dong, X.~Liang, X.~Shen, B.~Wang, H.~Lai, J.~Zhu, Z.~Hu, and J.~Yin,
  ``{Towards Multi-Pose Guided Virtual Try-on Network},'' in
  \emph{International Conference on Computer Vision (ICCV)}, 2019, pp.
  9026--9035.

\end{thebibliography}

\newpage

\title{FICE: Text--Conditioned Fashion Image Editing With Guided GAN Inversion\\Supplementary Material}

\maketitle

\begin{abstract}
In the main part of the paper, we reported several results to highlight the capabilities of FICE. In this \textit{Supplementary material}, we present additional details and experiments to further explore the characteristics of FICE, including: $(i)$ details on the hyperparameter settings for the competing techniques (pSp, E4e, ReStyle and HyperStyle) used in the main part of the paper, $(ii)$ investigations into alternative latent code initialization schemes (with style--mixing), and $(iii)$ additional results on the MPV image dataset. 

\end{abstract}


\section{Hyperparameter Settings}

In the main part  of the paper, we considered several baseline GAN inversion techniques combined with StyleCLIP to compare FICE against. These included {pSp}~\cite{psp}, {E4e}~\cite{e4e}, {ReStyle}~\cite{restyle} with the pSp and E4e backbones, and {HyperStyle}~\cite{hyperstyle}. The optimal hyperparameter ($\alpha,\beta$) settings that were used with these methods are listed in Table~\ref{tab:hyperparameter_options} for completeness.  

\begin{table}[h]
    \centering
    \caption{\textbf{Best performing hyperparamer settings.} The best setting  for each model was determined based on the semantic-relevance score.}
    \resizebox{0.95\columnwidth}{!}{%
    \begin{tabular}{l|rrrr}
    \hline \hline
    \textbf{Model} && \textbf{Magnitude} ($\alpha$) && \textbf{Disentanglement} ($\beta$) \\
    \hline
        pSp \cite{psp}               && $4.0$ && $0.50$ \\
        e4e \cite{e4e}               && $4.0$ && $0.025$ \\ 
        ReStyle-pSp \cite{restyle}   && $7.5$ && $0.025$ \\
        ReStyle-e4e \cite{restyle}   && $7.5$ && $0.050$ \\
        HyperStyle \cite{hyperstyle} && $5.0$ && $0.050$ \\
         \hline \hline
    \end{tabular}
    }
    \label{tab:hyperparameter_options}
\end{table}

\section{Exploring Style Mixing for Code Initialization}
As demonstrated in the experiments in the main part of the paper, FICE generates competitive high-quality editing results when compared to state-of-the-art models from the literature. Nevertheless, we observe that in certain cases, the results generated by FICE are impacted by the characteristics of the input image. When the targeted semantics from the provided text description  differ significantly from the semantics already present in the input image (i.e. changing a plain white shirt to a dark coloured shirt), FICE can sometimes produce unsatisfying results. In this subsection, we, therefore, explore alternative ways of latent code initialization to mitigate this issue and synthesize images with better target semantics.

\begin{figure}[t]
	\centering
\includegraphics[width=\columnwidth]{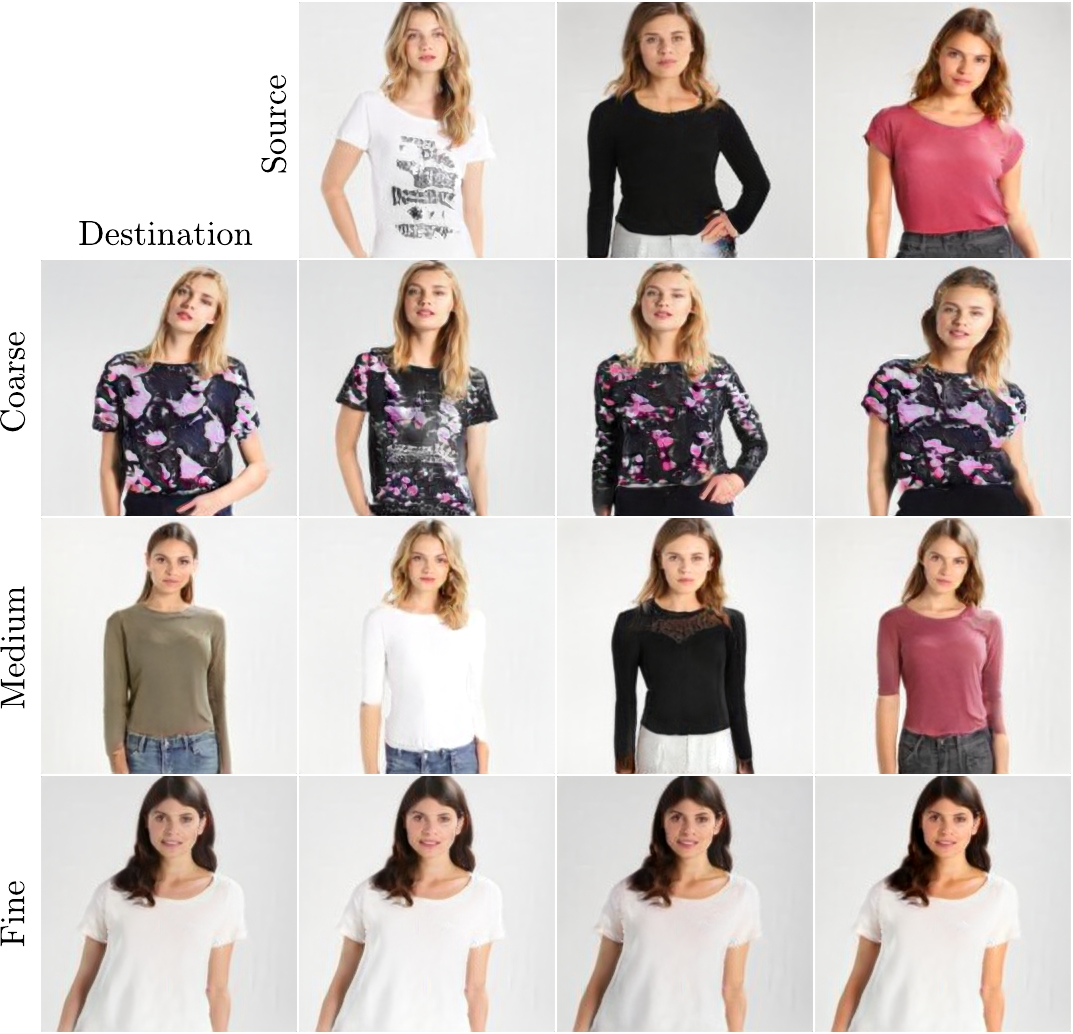}
	\caption{\textbf{Example results of style--mixing experiments.} We take part of the latent code from the Source image and use it to replace the corresponding part in the latent code of the Destination image. Only a certain subset of the original code of the Destination image is replaced, while the rest is preserved. 
	We observe that copying the \textit{coarse} subset (layers $1$ to $4$) causes the destination image to exhibit the pose of the source image. The \textit{medium} subset (layers $5$ to $8$) appears most suitable for our task, as it tends to replicate the clothing style of the source image, while preserving the pose of the destination image. Finally, copying the \textit{fine} subset (layers $9$ to $14$) mostly results in minor changes in the image tone without major impact on the clothing or pose of the Destination image.}
	\label{fig:app:mix_styles}
\end{figure}

\renewcommand{\iwidth}{.233\columnwidth}
\renewcommand{\im}[1]{\includegraphics[width=\iwidth]{figures/mix-styles/initialization_example/#1.jpg}}
\setlength{\tabcolsep}{2pt}

\begin{figure}
	\begin{tabular}{CCCC}
		\im{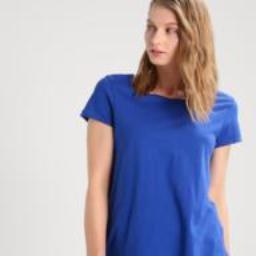} &\im{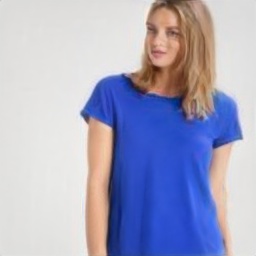} & \im{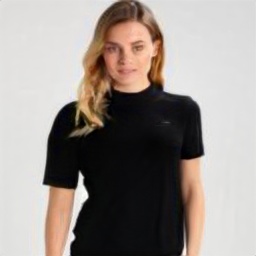} & \im{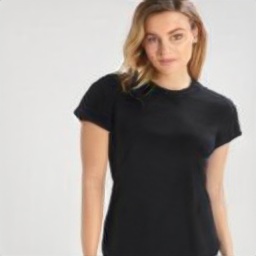} \\
		Input  &E4e Image & CLIP Prototype &Code Injection \\
	\end{tabular}
	\caption{\textbf{Latent code initialization with style mixing (injection).} The examples show the initialization process for an input image and the following text description \textit{"Short sleeve antimicrobial merino wool-blend t-shirt in black"}. The input image is processed with the E4e model to obtain a latent code that corresponds to the input image (2nd column). Based on the text description, we identify a suitable CLIP prototype code (3rd column), which is injected into the computed E4e code, resulting in an image (4th column) with  similar pose to the input image and clothes resembling the identified CLIP prototype.}
	\label{fig:app:initialization}
\end{figure}

We note again that FICE operates in the extended latent vector space $\mathcal{W}^+$ of the pretrained StyleGAN generator. The complete latent code $w \in \mathcal{W}^+$ of a given input image $I$, therefore, consists of several individual latent codes, each impacting an individual convolutional layer in the StyleGAN generator network $G$.
To better understand the semantics, encoded in different subsets of the overall latent code, we conduct style mixing experiments in this section. Style mixing refers to injecting a latent-code subset into another latent code. Similarly, as in \cite{stylegan1}, we do so for coarse, medium and fine subsets of the latent code $w=\{w_l\}_{l=1}^L$, (with $L=14$ for our implementation of StyleGANv2), where the coarse subset corresponds to $l \in \{1, ..., 4\}$, medium to $l \in \{5, ...,8\}$, and fine
to $l \in \{9, ..., 14\}$ layers. A few example results of style--mixing experiments are presented in Figure~\ref{fig:app:mix_styles}. We observe that copying part of the latent code that corresponds to the medium subset (layers $5$ to $8$) results in images with roughly the same pose as the original (destination) image, while inheriting the (approximate) clothing style of the source image. 

\setlength{\tabcolsep}{2pt}
\renewcommand{\arraystretch}{1.2}
\begin{table}[t]
	\centering
	\caption{\textbf{Quantitative results.} The style--mixing (code injection) initialization procedure improves the semantic-relevance score, but degrades other performance indicators. }
	\resizebox{0.99\columnwidth}{!}{%
	\begin{tabular}{lc|cccccccr}
        \hline
        \hline
             \textbf{Initialization}                        &&& \textbf{Semantics} ($\uparrow$) && \textbf{Identity sim.} ($\uparrow$) && \textbf{IoU} ($\uparrow$)   & &\textbf{FID} ($\downarrow$)     \\
        \hline
E4e init. (FICE)                &&& $0.446$         && $\mathbf{0.926}$ && $\mathbf{0.949}$ & &$\mathbf{60.96}$ \\
Injection init. &&& $\mathbf{0.468}$ &&           $0.912$ &&         $0.931$          && $84.03$  \\
        \hline
        \hline 
	\end{tabular}
	\label{tab:app:code_injection}
	}
\end{table}

\renewcommand{\iwidth}{.22\columnwidth}
\renewcommand{\twidth}{.03\columnwidth}
\renewcommand{\im}[1]{\includegraphics[width=\iwidth]{figures/mix-styles/result_comparison/#1.jpg}}
\renewcommand{\imrow}[1]{\im{init_e4e/#1/035} & \im{init_inject/#1/035} & & \im{init_e4e/#1/089} & \im{init_inject/#1/089} }

\begin{figure}[t]
    \centering
	\begin{tabular}{CCDCC}
   \multicolumn{2}{c}{\im{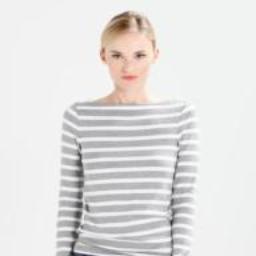}}  &  & \multicolumn{2}{c}{\im{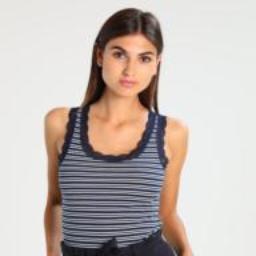}} \\ 
		\multicolumn{5}{c}{Input} \\ 
		\hdashline  \\
		
		\imrow{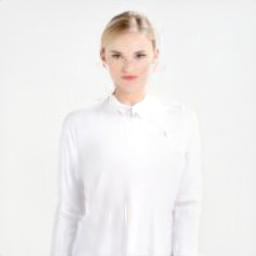} \\
		\multicolumn{5}{c}{\textit{Long sleeve cotton sateen shirt in white.}} \\
		\imrow{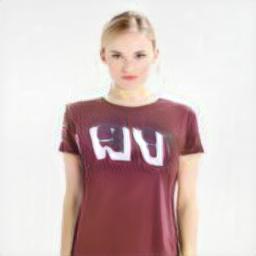} \\
		\multicolumn{5}{c}{\textit{Short sleeve jersey t-shirt in burgundy.}} \\
		\imrow{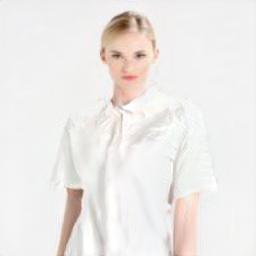} \\
		\multicolumn{5}{c}{\textit{Short sleeve semi-sheer silk shirt in white.}} \\
		FICE & Injection & & FICE & Injection
	\end{tabular}
	\caption{\textbf{Comparison of latent-code initialization procedures.} The figure shows example results when initializing the latent codes needed by FICE wither with the (style--mixing based) code injection and the vanilla E4e initialization used in the main paper. The first row shows the input images. The rows below show a comparison of the results when either initializing with E4e encoder (FICE) or when initializing with the code injection. 
	We observe that the code-injection technique for images with certain characteristics produces better results than FICE. Specifically, the code injection technique tends to facilitate better edits with respect to sleeve length and independence of the initial clothing characteristics -- see the results corresponding to the stripe-pattern (left column).}
	\label{fig:app:code_injection}
\end{figure}
	
\renewcommand{\iwidth}{.16\textwidth}
\renewcommand{\im}[1]{\includegraphics[width=\iwidth]{figures/MPV_results/#1.jpg}}

\renewcommand{\imrow}[1]{\im{#1} & \im{img-#1/096} & \im{img-#1/102} & \im{img-#1/200} & \im{img-#1/209} & \im{img-#1/228}}

\begin{figure*}[t]
	\centering
	\begin{tabular}{CCCCCC}
\imrow{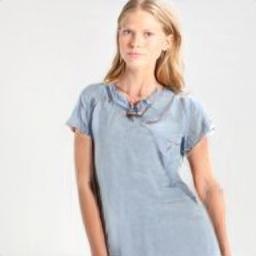} \\
\imrow{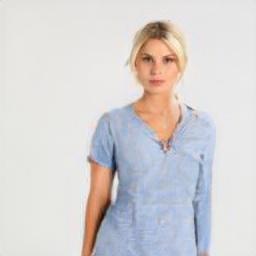} \\
\imrow{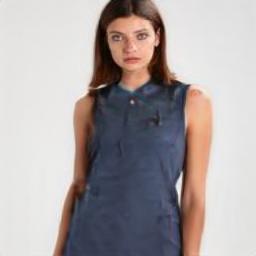} \\
\imrow{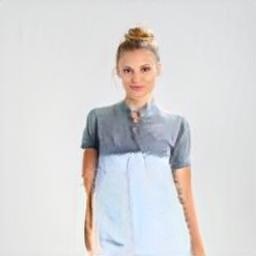} \\
\imrow{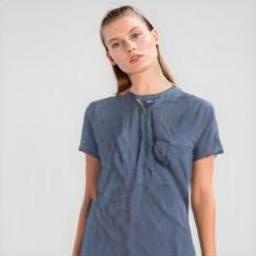} \\
Input & \scriptsize \textit{Short sleeve chambray shirt-dress in blue.}	& \scriptsize \textit{Short sleeve jersey t-shirt in burgundy.}  & \scriptsize \textit{Short sleeve cotton jersey t-shirt in dahlia pink.} & \scriptsize \textit{Short sleeve cotton piqu\'e polo in grey.} & \scriptsize \textit{Long sleeve silk crepe de chine shirt in sprint green.} \\
	\end{tabular}
    	\caption{\textbf{Example results generated by FICE for various text descriptions for the MPV dataset.} We again observe that the results preserve the pose and identity of the subjects, as well as other image characteristics, such as scene illumination.} 
	\label{fig:app:MPV_results}
\end{figure*}

The above observations motivate us to experiment with a different latent-code initialization procedure than used in the main part of the paper,  where the coarse and fine subsets of the latent code are related to the input image, while the medium subsets exhibits visual semantics that correspond to provided text description $t$. In order to obtain the latent code that best corresponds to the given text description, we use a sampling approach. Specifically, we generate $N$ (complete) latent codes $w^{(i)}$ that serve as the \textit{prototypes} for our initialization procedure and are drawn randomly from different parts of the GAN latent space. Based on the sampled prototypes, we then generate the corresponding CLIP image embeddings $e^I_i = C^i(G(w^{(i)})) \in \mathcal{R}^{d_{clip} \times 1}$. Finally, we construct $N=100,000$ $(w^{(i)}, e^I_i)$ pairs and store them for later processing. 

When editing an image given the text description $t$, we process the text with the CLIP text encoder $C^t$ to obtain the text embedding $e^T = C^t(t) \in \mathcal{R}^{d_{clip} \times 1}$ and compute all $N$ similarities: 
\begin{equation} \label{eq:clip-similarity}
	S_i(e^I_i, e^T) = \cos (e^I_i, e^T),
\end{equation}
where $i\in\{1,\cdots,N\}$.
The target prototype $w^{(i^*)}$, providing the medium latent code subset is then selected based on the maximum similarity, i.e., $i^* = \argmax_i\{S_i\}$. Finally, to obtain the coarse and fine latent-code subsets that best match the input image, we again leverage the E4e encoder to predict the extended latent code of the input image before and inject it with the medium latent code subset of the selected prototype $w^{(i^*)}$. The complete process is visualized in Fig.~\ref{fig:app:initialization}.


We evaluate the original initialization procedure and the style--mixing (with prototypes) initialization procedure quantitatively and qualitatively. In Table~\ref{tab:app:code_injection}, we show the results with respect to our performance indicators. We observe that the semantic-relevance score does increase, suggesting that the semantics, expressed in the text descriptions, are now better integrated into the edited images (on average). However, all other performance indicators exhibit a slight degradation, most obviously, the FID score. Nevertheless, there are several positive aspects of such an initialization technique, as we show in Fig.~\ref{fig:app:code_injection}. Note how the alternative initialization (marked \textit{injection}) scheme allows us to convincingly infuse semantics that differ considerably from the original image. With the original initialization process this is not always the case.

\section{MPV Dataset}
We explore the generalization capabilities of FICE by investigating text-conditioned image editing on the MPV image dataset~\cite{dong2019towards}. We process the images with the same preprocessing operations as used for VITON dataset, first cropping the bottom part of the image to $192 \times 192$ px, then resizing the image to $256 \times 256$ px. For the experiments, we again create various image-text combinations, where the text descriptions stem from the Fashion-Gen dataset. Qualitative example results are shown in Fig.~\ref{fig:app:MPV_results}. Note how FICE is again able to convincingly infuse the semantics from the  text descriptions into the MPV images, while preserving the pose and identity of the subjects in the input images.


\section*{Acknowledgements}

Supported in parts by the Slovenian Research Agency ARRS through the Research Programme P2-0250(B) Metrology and Biometric System, the ARRS Project J2-2501(A) DeepBeauty and the ARRS junior researcher program.

\end{document}